\documentclass[11pt]{article}

\usepackage[T1]{fontenc}
\usepackage[margin=0.9in]{geometry}
\usepackage{amsmath,amssymb}
\usepackage{graphicx}
\usepackage{authblk}
\usepackage{xcolor}
\usepackage{tikz}
\usetikzlibrary{positioning,arrows.meta,fit,backgrounds}
\usepackage{natbib}
\usepackage{algorithm}
\usepackage{algorithmic}
\usepackage{multirow}
\usepackage{makecell}
\newcommand{\appref}[1]{Appendix~\ref{#1}}
\usepackage{amsthm}
\usepackage{booktabs}
\newtheorem{remark}{Remark}
\usepackage[font=small]{caption}
\usepackage{academicons}
\usepackage{xcolor}
\usepackage[hidelinks]{hyperref}

\title{Quantifying structural stability of transparent models \\for general insurance pricing}

\author{Félix Côté and Marie-Pier Côté}
\affil{École d'actuariat, Université Laval, Canada}

\date{November 12, 2026}

\begin{document}

\maketitle

\begin{abstract}
\leftskip=-0.5em \rightskip=-0.5em
The short duration of general insurance contracts leads to rapid distributional shifts in the data, warranting regular model updates. This process may yield a post-update model that differs considerably from the pre-update model in both predictions and structure. Structural stability is upstream of predictive stability, yet existing structural stability metrics are mostly restricted to simple decision trees. Thus, we first introduce the notion of functional transparency, characterising models with a prediction function that can be expressed as a sum of univariate functions for main effects and bivariate functions for pairwise interactions. Then, we propose the Mean Squared Shape function distance, a metric to quantify the structural stability of functionally transparent models, which are particularly relevant in actuarial science. Using three public auto insurance datasets, we present a benchmark study to demonstrate the usefulness of our metric in identifying a stable and accurate transparent model. Generalised linear models with centred numerical covariates are stable, and explainable boosting machines (EBMs) offer a compromise between stability and performance. We also find that stabilisation strategies, such as sequential modelling and fixed hyperparameters, can substantially improve stability for both EBM and XGBoost models, with negligible impact on performance.

\end{abstract}


\section{Introduction}
\label{sec1}

In general insurance, the short duration of contracts induces a fast-paced environment that compels frequent model retraining to leverage the information brought by newly collected data. During these updates, the model class typically remains unchanged, while hyperparameters are selected anew and parameters are re-estimated based on all available training data. Updating the model allows us to keep pace with shifts in the data distribution \citep{liu2022model} and to reduce the variance of the estimators by incorporating additional data. However, in doing so, we must manage retraining costs and ensure smooth model transitions over time \citep{bertsimas2025stablemachinelearningmodel}. A framework to consider the trade-off between the expected return of adaptation and the cost of pursuing updates \citep{vzliobaite2015towards}, or automated systems designed to identify cost-efficient retraining opportunities \citep{mahadevan2023costeffectiveretrainingmachinelearning}, 
address these issues.

When using machine learning for high-stakes 
applications such as criminal justice, healthcare, computer vision or insurance, new retraining-related challenges emerge. In particular, updating a model can cause considerable changes in both its structure and predictions. As highlighted by \citet{glikson2020human}, such changes may lead domain experts to mistrust predictions due to inconsistent interpretations across model iterations. In some cases, this can result in outright rejection of the model \citep{longoni2019resistance}. One way to overcome these problems is to evaluate and monitor model stability throughout updates; this requires a precise definition of stability.

The most widespread definition of model stability originates from \citet{breiman1996heuristics}, who describes a stable model as one that is insignificantly affected by small changes in the data. Predictive stability, the earliest approach to quantify stability, is closely linked to sensitivity analysis: it measures how much the predictions of a model (the outputs) are affected by alterations to the training data (the inputs). Advantages of this approach are its strong relationship with generalisation error \citep{bousquet2002stability} and its model-agnostic design, which enables its use with black box models. Notable work on predictive stability includes the agreement metric (\citealp{schaffer1992empirical}; \citealp{turney1995bias}), analyses of random forest stability \citep{wang2023stability} and stability guarantees induced by bagging methods \citep{soloff2024bagging}. However, predictive stability metrics are restricted to either model-level or observation-level assessments and preclude exact feature-level assessments. Methods based on post-hoc explanation stability \citep[see, e.g.,][]{zhao2026explainable} are restrained to approximate feature-level analysis. Without exact feature-wise contributions to the global or local model instability, the predictive approach is limited for comprehensively assessing stability and for guiding effective stabilisation strategies. As hinted by \citet{breiman1996heuristics}, although predictive stability offers desirable properties, it is ``not a panacea''. Indeed, different model structures can produce identical predictions. To understand predictive stability and build confidence in deployed models as evolving representations of the world, we should leverage the model's structure to quantify stability.

Structural stability, which measures how much the structure of a model is affected by alterations to the training data, can be viewed as a more upstream notion than predictive stability; controlling the structure that governs predictions is a more demanding requirement than controlling the individual predictions themselves. Due to their transparent graphical representation, decision trees are extensively studied in terms of structural stability (see, e.g., \citealp{last2002improving}; \citealp{dwyer2007decision}; \citealp{zimmermann2008ensemble}; \citealp{briand2009similarity}; \citealp{mirzamomen2017framework}; \citealp{bertsimas2023improvingstabilitydecisiontree}). Indeed, for any given observation, we can trace the path from the root of the tree to the corresponding leaf node, providing a clear understanding of how each feature contributes to the prediction, which enables quantification of structural stability. However, the simplicity of decision trees hinders them from adequately modelling the complexities of insurance datasets \citep[see, e.g.,][]{henckaerts2021boosting}. Research on structural stability beyond decision trees remains scarce, as does work on a general definition of structure needed to advance this work.

Although actuarial pricing traditionally relied on generalised linear models (GLMs) (see, e.g., \citealp{Jong/Heller:2008}; \citealp{frees2014predictive}; \citealp{wuthrich2026ai}), recent works show that complex algorithms, such as gradient boosted decision trees \citep{friedman2001greedy}, can outperform them in terms of predictive power (\citealp{Yang/etal:2018}; \citealp{henckaerts2021boosting}). Gradient boosting leads to a fine segmentation of risks, aligning premiums with expected costs, which is desirable from an actuarial fairness perspective. Nonetheless, such black box algorithms exhibit two major drawbacks for insurance pricing. First, their interpretation relies on post-hoc explanation methods that are vulnerable to manipulation \citep[see, e.g.,][]{xin2024trustinterpretationsmachinelearning}. Second, because structural stability cannot be quantified, reliance on predictive stability is required to manage variations in individual predictions during retraining. This calls for metrics to quantify structural stability of complex transparent models.

In this paper, we first introduce the notion of functional transparency, according to which the prediction can be expressed as a sum of univariate functions for main effects and bivariate functions for pairwise interactions. It allows easy visualisation and interpretation of the feature effects. GLMs, generalised additive models (GAMs), and the explainable boosting machines (EBMs) of \citet{lou2013accurate} are all functionally transparent. We propose a structural stability metric for such transparent models, called the Mean Squared Shape function distance (MSS). Modelling frequency, severity, and loss cost on three public auto insurance datasets, our benchmark study illustrates the usefulness of the MSS in identifying a transparent model that is both stable and high-performing. We investigate strategies such as sequential modelling and the use of fixed hyperparameters or interaction terms. We find that these strategies can stabilise models while having minimal impact on predictive performance.

The remainder of this paper is organised as follows. In Section~\ref{sec:methodology}, we briefly outline two common model retraining approaches and describe how retraining is conducted for our case studies. In Section~\ref{sec:background_stability}, we introduce the notion of functional transparency. In Section~\ref{sec:stability_metrics}, we propose the MSS, a data-dependent structural distance metric tailored to functionally transparent models relevant to high-stakes decision-making. We also recall existing predictive stability and performance metrics that we monitor. In Section~\ref{sec:control}, we present the two stabilisation strategies considered in the benchmark study of Section~\ref{sec:case_study}. We conclude in Section~\ref{sec:conclusion} and provide additional results in appendices.

\section{Model update via batch learning}
\label{sec:methodology}

Two typical approaches for updating models are online learning and batch learning. With online learning, we incrementally update the model as new data become available continuously, which requires extremely frequent updates. Even though computationally efficient methods tackle this issue \citep{gama2014survey}, they often use partial data, which leads to greedy solutions with possibly reduced generalisation capacity. Furthermore, frequent changes to models in production might be difficult to implement in a highly-regulated industry such as insurance. Batch learning, on the other hand, consists of periodic retraining of the model from scratch using the entire updated dataset, ensuring globally optimal solutions. As shown in \citet{zanotti_retraining_2025}, batch learning balances predictive performance and efficiency. In fact, reducing the retraining frequency preserves forecast accuracy and lessens computational and implementation costs. \citet{bertsimas2025stablemachinelearningmodel} also show that periodic (as opposed to continuous) data updates enhance interpretability and help mitigate model instability. For these reasons, we only consider periodic retraining in this work. However, our structural stability metric does not rely on the specific setting considered here; it could also be monitored for online learning. In the remainder of this section, we describe the notation and our model retraining scheme.

For observation $i \in \mathcal{I}$, let $y_i \in \mathcal{Y}$ be the response variable, let $\mathbf{x}_i = (x_{ij})_{j \in \mathcal{J}}$ be the associated $p$-dimensional vector of covariates valued in $\mathcal{X} = \mathcal{X}_1 \times \dots \times \mathcal{X}_p$ where $\mathcal{J} = \{1, \dots, p\}$, and let $t_i \in \mathbb{R}^+$ be the time at which observation $i$ was collected. In general, we assume the datasets are of the form $\mathcal{D}=\{( y_i, \mathbf{x}_i, t_i)\}_{i \in \mathcal{I}}$. We take the cardinality of the dataset $|\mathcal{D}|$ as the cardinality of its index set $|\mathcal{I}|$, that is, the number of independent observations it contains.

Suppose we start by training and evaluating an initial model (the pre-update model), based on a pre-update dataset $\mathcal{D}^{\text{pre}}$ formed by all data points observed before a given time $T \in \mathbb{R}^+$. This dataset is partitioned into a training set $\mathcal{D}^{\text{pre}}_{\text{train}}$ and a test set $\mathcal{D}^{\text{pre}}_{\text{test}}$ by sampling based on a train proportion $\alpha \in (0,1)$. 
We apply the same proportion~$\alpha$ to partition the new batch of data that is available at time $T^{*} > T$, called the update dataset $\mathcal{D}^{\text{up}}$, into $\mathcal{D}^{\text{up}}_{\text{train}}$ and $\mathcal{D}^{\text{up}}_{\text{test}}$. The post-update training and test datasets $\mathcal{D}^{\text{post}}_{\text{train}}$ and $\mathcal{D}^{\text{post}}_{\text{test}}$ are then constructed by merging the corresponding folds
$$
\mathcal{D}^{\text{post}}_{\text{train}} = \mathcal{D}^{\text{pre}}_{\text{train}} \cup \mathcal{D}^{\text{up}}_{\text{train}} \quad \text{and} \quad \mathcal{D}^{\text{post}}_{\text{test}} = \mathcal{D}^{\text{pre}}_{\text{test}} \cup \mathcal{D}^{\text{up}}_{\text{test}}.
$$ 
The model is then retrained from scratch on $\mathcal{D}^{\text{post}}_{\text{train}}$, yielding the post-update model. For example, retraining a Poisson GLM with log link retains the Poisson and log-linear assumptions; however, we perform variable and interaction selection, and estimate parameters anew on the post-update training data. We wish to study the stability between the pre-update and the post-update models. 

\begin{figure}[b!]
    \centering
    \makebox[\textwidth][c]{%
      \includegraphics[width = 0.95\textwidth]{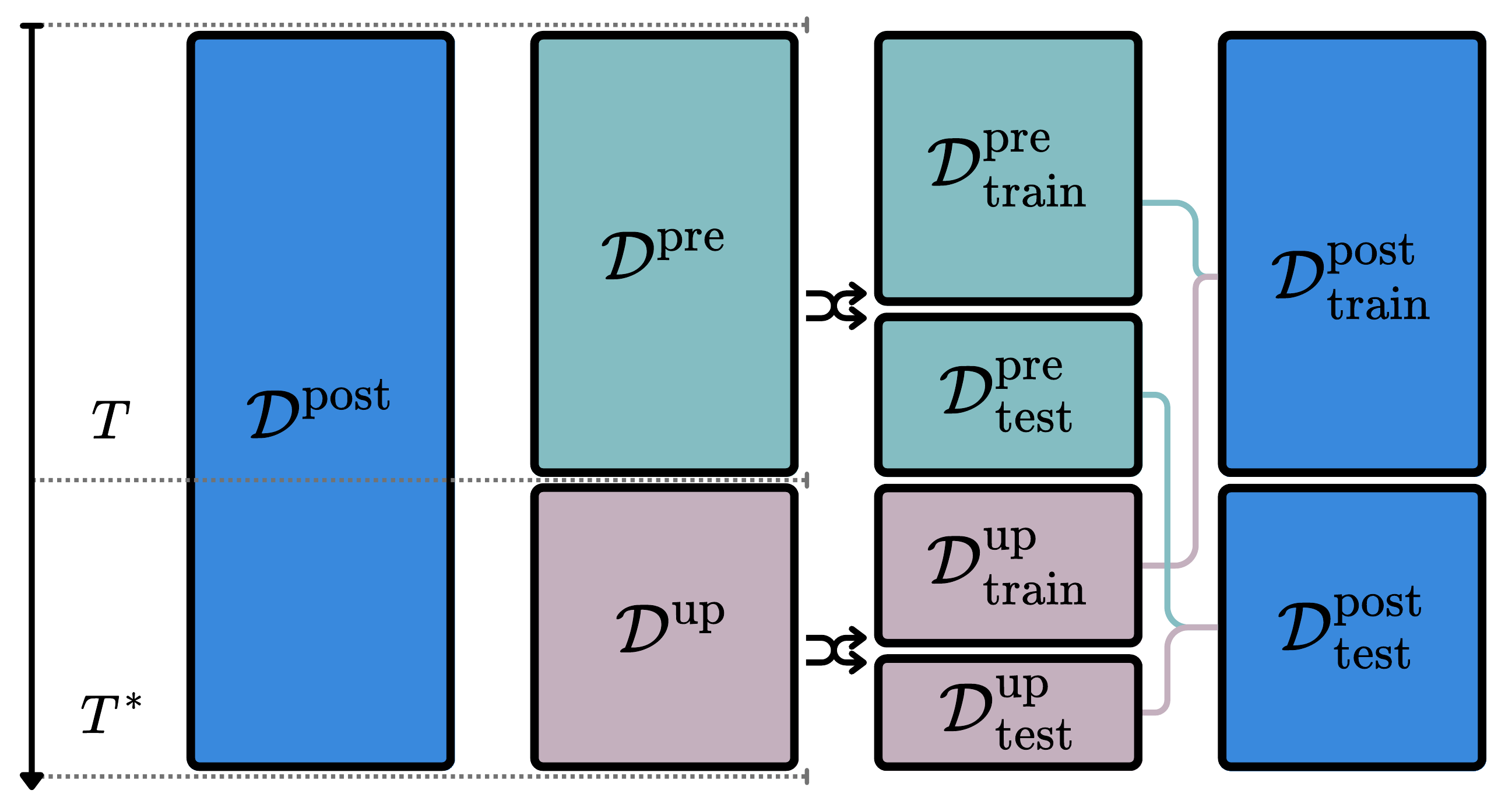}}
    \caption{Construction of the pre-update and post-update training and test datasets using a time stamp.}
    \label{figs:partitionning_temp}
\end{figure}

Note that the post-update dataset, consisting of all available data at time $T^{*}$, is 
$$
\mathcal{D}^{\text{post}} = \mathcal{D}^{\text{post}}_{\text{train}} \cup \mathcal{D}^{\text{post}}_{\text{test}} = \mathcal{D}^{\text{pre}} \cup \mathcal{D}^{\text{up}}.
$$
Thus, we have $\mathcal{D}^{\text{pre}} \subseteq \mathcal{D}^{\text{post}}$ for our case study, although this is not a requirement for our metric. In Algorithm~\ref{alg:data-split}, we describe how to partition and merge subsets of a single complete dataset $\mathcal{D}^{\text{post}}$ to construct the pre-update and post-update training and test datasets, required to emulate a model update. Figure~\ref{figs:partitionning_temp} illustrates the procedure. The shuffle icons indicate that the observations are permuted before the train--test split. 
This avoids a temporal dependence into the split. Note that Figure~\ref{figs:partitionning_temp} represents cross-sectional data for simplicity. With longitudinal data, observations relating to a policyholder must not appear in both the training and test sets, requiring extra care in step 6 of Algorithm~\ref{alg:data-split} to make sure the train-test splits are consistent for the pre-update and update datasets. When there is no time stamp in the complete dataset, Algorithm~\ref{alg:data-split-variation} in \appref{app:algo2} can be used.

\begin{algorithm}[b!]
    \caption{Construction of the pre- and post-update training and test datasets with time stamps.}
    \label{alg:data-split}
    \renewcommand{\algorithmicrequire}{\textbf{Input:}}
    \renewcommand{\algorithmicensure}{\textbf{Output:}}
    \renewcommand{\algorithmicreturn}{\textbf{Return}}
    
    \begin{algorithmic}[1]

    \REQUIRE Complete dataset $\mathcal{D}^{\text{post}}$, previous update time $T \in \mathbb{R}^+$, train proportion $\alpha \in (0,1)$.\\
    
    \ENSURE Pre-update and post-update training and test datasets.

    \STATE Set $n^{\text{post}} := |\mathcal{D}^{\text{post}}|$.
    
    \STATE Split $\mathcal{D}^{\text{post}}$ into the pre-update dataset 
    $$
    \mathcal{D}^{\text{pre}}=\{(y_i, \mathbf{x}_i,, t_i): t_i \leq T, i = 1, \dots,n^{\text{post}}\}$$ and the update dataset $$\mathcal{D}^{\text{up}}=\{(y_i, \mathbf{x}_i, t_i): t_i > T, i = 1, \dots,n^{\text{post}}\}.
    $$

    \STATE Set $n^{\text{pre}} := |\mathcal{D}^{\text{pre}}|$ and $n^{\text{up}} := |\mathcal{D}^{\text{up}}|$.

    \STATE Sample the pre-update training dataset $\mathcal{D}^{\text{pre}}_{\text{train}}$ by drawing without replacement $\lfloor \alpha n^{\text{pre}}\rfloor$ observations from $\mathcal{D}^{\text{pre}}$.
    % Recover
    \STATE Extract the pre-update test dataset $\mathcal{D}^{\text{pre}}_{\text{test}} = \mathcal{D}^{\text{pre}}\backslash \mathcal{D}^{\text{pre}}_{\text{train}}$. 

    \STATE Sample the update training dataset $\mathcal{D}^{\text{up}}_{\text{train}}$ by drawing without replacement $\lfloor \alpha n^{\text{\text{up}}}\rfloor$ observations from $\mathcal{D}^{\text{\text{up}}}$.

    \STATE Extract the update test dataset $\mathcal{D}^{\text{\text{up}}}_{\text{test}} = \mathcal{D}^{\text{\text{up}}}\backslash \mathcal{D}^{\text{\text{up}}}_{\text{train}}$. 

    \STATE Assemble the post-update training and test datasets $\mathcal{D}^{\text{post}}_{\text{train}} = \mathcal{D}^{\text{pre}}_{\text{train}} \cup \mathcal{D}^{\text{\text{up}}}_{\text{train}}$ and $\mathcal{D}^{\text{post}}_{\text{test}} = \mathcal{D}^{\text{pre}}_{\text{test}} \cup \mathcal{D}^{\text{\text{up}}}_{\text{test}}$.

    \RETURN $\mathcal{D}^{\text{pre}}_{\text{train}}$, $\mathcal{D}^{\text{post}}_{\text{train}}$, $\mathcal{D}^{\text{pre}}_{\text{test}}$, $\mathcal{D}^{\text{post}}_{\text{test}}$.
    \end{algorithmic}

\end{algorithm}

\section{Functional transparency}
\label{sec:background_stability}

As detailed by \citet{lipton2018mythos}, the notion of inherent interpretability, or transparency, takes many forms. For example, decomposability requires that the input features, the model's parameters, and the computations performed are understandable; simulatability demands that a human can take the input data along with the model's parameters and, within a reasonable time, reproduce each computation leading to the final prediction; algorithmic transparency necessitates that the learning algorithm yields an optimal unique solution. The relevance of each form depends on the use case. For insurance applications, none of these forms is adequate on its own. Decomposability is necessary yet insufficient. We need a stronger version, in which we exactly understand the contribution of each input on the prediction. In this section, we thus define a new form, that we call functional transparency.

Functional transparency for a model requires that its 
prediction function is a sum of univariate functions for main effects and bivariate functions for pairwise interactions. This definition of model structure allows to visualise and interpret the exact contribution of each term (main effect or interaction) to the prediction function $\hat{\mathcal{F}}$. Such models are limited to pairwise interactions, because higher-order terms require abstract visualisation tools to be understood. More specifically, the contribution of a term to the prediction function of a model is described by a univariate shape function $\hat{f}_{j}$ in the case of a main effect $j \in \mathcal{J}$, or by a bivariate shape function $\hat{f}_{\boldsymbol{\ell}}$ in the case of a pairwise interaction~$\boldsymbol{\ell}\in\mathcal{L} =\{(\ell_1,\ell_2): \ell_1,\ell_2\in\mathcal{J}, \ell_1 < \ell_2\}$. Model classes adhering to this form of transparency, such as GLMs and GAMs \citep{hastie1986generalized}, share a common additive structure. Using the superscript $a \in \{\text{pre}, \text{post}\}$ to indicate whether the model associated with a component is trained on the pre- or post-update dataset, and $g: \mathcal{Y} \to \mathbb{R}$ to denote a differentiable and strictly monotonic link function, the prediction function of functionally transparent models can be expressed, for $\mathbf{x}_i \in \mathcal{X}$, as
\begin{align}
    \hat{\mathcal{F}}^a(\mathbf{x}_i) = g^{-1} \bigg\{\underbrace{\hat{\beta}^a_{0}}_{\text{intercept}} + \underbrace{\sum_{j \in \mathcal{J}} \hat{f}^a_{j} (x_{ij})}_{\text{main effects}} + \underbrace{\sum_{\boldsymbol{\ell} \in \mathcal{L}} \hat{f}^a_{\boldsymbol{\ell}} (\mathbf{x}_{i \boldsymbol{\ell}})}_{\text{pairwise interactions}}\bigg\},
    \label{eq:additive_form}
\end{align}
where $\mathbf{x}_{i \boldsymbol{\ell}} = (x_{ij})_{j \in \boldsymbol{\ell}}$ are the features valued in $\mathcal{X}_{\boldsymbol{\ell}}$ associated to interaction $\boldsymbol{\ell}$ for observation $i$. If a covariate $j$ or an interaction $\boldsymbol{\ell}$ is not selected in model $a$, we set the associated shape function to zero, that is, $\hat{f}_j^a(x_{ij}) :=0$ or $\hat{f}_{\boldsymbol{\ell}}^a(\mathbf{x}_{i\boldsymbol{\ell}}) :=0$, respectively.
 
Two approaches, based on GLMs and GAMs, exist for designing more flexible models that satisfy functional transparency. The first aims to distill knowledge from black box models and integrate it into a GLM within a data-driven framework. \citet{henckaerts2022stakes} follow this approach with a Model-Agnostic Interpretable Data-driven suRRogates (maidrr), a procedure that leverages the partial dependence plots \citep{friedman2001greedy}, computed from a black box algorithm, to automatically preprocess explanatory variables provided as inputs to a categorical GLM. Similarly, \citet{lindholm2024black} propose a guided categorical GLM. However, instead of discretising features using the dynamic programming algorithm of \citet{wang2011ckmeans} as preprocessing, they use the Classification and Regression Trees algorithm \citep{breiman1984classification}. They also add a lasso-regularisation \citep[see, e.g.,][]{hastie2015statistical} when fitting the final guided categorical GLM.

The second approach aims to design, within a GAM framework, shape functions that  are more expressive than typical splines. \citet{lou2012intelligible} present a large-scale empirical comparison of existing methods for learning GAMs; they compare spline and tree-based ensemble methods for shape functions, and penalised iteratively reweighted least squares \citep{wood2017generalized}, gradient boosting \citep{friedman2001greedy}, and backfitting \citep{hastie1986generalized} as learning algorithms. Their experiments show that combining bagged-boosted shallow decision trees for the shape functions with gradient boosting as the learning algorithm substantially improves accuracy over other combinations for both regression and classification problems. 

Another modelling choice, beyond the specification of the shape functions and the learning algorithm, is the method to select interactions. While interactions are typically added manually based on prior knowledge, more advanced methods enable their data-driven selection. Existing techniques include ANOVA (\citealp{hastie1986generalized}; \citealp{wood2017generalized}), GUIDE \citep{loh2002regression}, Grove \citep{sorokina2008detecting}, and thresholding of the Friedman's $H$ statistic \citep{friedman2008predictive}, as in \citet{henckaerts2022stakes}. \citet{lou2013accurate} introduced a competing method called FAST; they show that, although Grove remains the most accurate method currently available, FAST is considerably more efficient, making it scalable to large datasets while maintaining high accuracy. Building the best GAM first and then using an algorithm to automatically select interactions yields a model class known as Generalised Additive Model plus Interactions ($\text{GA}^2\text{M}$), as coined by \citet{lou2013accurate}. 

By adopting what \citet{lou2012intelligible, lou2013accurate} identify as the best choices for the shape functions type, the learning algorithm, and the interaction selection method, we obtain an EBM, as studied by \citet{caruana2015intelligible, nori2019interpretmlunifiedframeworkmachine}. In other words, an EBM is a GAM that relies on bagged ensemble of boosted shallow decision trees for the shape functions, gradient boosting for the learning algorithm, and FAST for interaction selection within a $\text{GA}^2\text{M}$ framework. \citet{chevalier2025point} and \citet{krupova2026explainableboostingmachinepredicting} show that EBM is a state-of-the-art transparent model in terms of predictive performance on insurance data, achieving performance comparable to (or slightly better than) the seminal black box XGBoost of \citet{chen2016xgboost}.

Functionally transparent models share a common structural representation because of the generalised additive form in Equation~\eqref{eq:additive_form}. This understanding of how each feature contributes to the prediction suggests that their structural stability can, in principle, be quantified. To the best of our knowledge, however, this has not yet been formally addressed; we close this gap in Section~\ref{sec:stability_metrics}. 

\section{Evaluation metrics}
\label{sec:stability_metrics}

A functionally transparent model that is structurally stable would preserve similar shape functions under small perturbations of the training data. Consequently, a structural stability metric should measure the distance between shape functions. Guided by this idea, we introduce our structural stability metric in Section~\ref{sec:sda}. We then present, in Section~\ref{sec:existing_metrics}, the existing predictive stability and performance metrics that we use in the case studies.

\subsection{Structural stability metric for functionally transparent models}
\label{sec:sda}

Assume for a moment that the covariates of a dataset~$\mathcal{D}$ are random; let $\mathrm{F}_{X_j}$ denote the marginal distribution of $X_j$ for $j\in \mathcal{J}$, and $\mathrm{F}_{\mathbf{X}_{\boldsymbol{\ell}}}$ denote the joint distribution of $\mathbf{X}_{\boldsymbol{\ell}}$ for $\boldsymbol{\ell}\in\mathcal{L}$. Then, a natural way to quantify the structural stability of functionally transparent models is through what we call the Expected Squared Shape function distance (ESS) between the pre- and post-update models (respectively noted $\hat{\mathcal{F}}^{\text{pre}}$ and $\hat{\mathcal{F}}^{\text{post}}$), that we define as
\begin{align*}
    \text{ESS}\big(\hat{\mathcal{F}}^{\text{pre}}, \hat{\mathcal{F}}^{\text{post}};& \mathcal{D}\big) = \sum_{j \in \mathcal{J}} \mathbb{E}_{X_j}\bigg[\big\{\hat{f}^{\text{post}}_{j}(X_{j}) - \hat{f}^{\text{pre}}_{j}(X_{j})\big\}^2\bigg] + 
    \sum_{\boldsymbol{\ell} \in \mathcal{L}} \mathbb{E}_{\mathbf{X}_{\boldsymbol{\ell}}}\bigg[\big\{\hat{f}^{\text{post}}_{\boldsymbol{\ell}}(\mathbf{X}_{\boldsymbol{\ell}}) - \hat{f}^{\text{pre}}_{\boldsymbol{\ell}}(\mathbf{X}_{\boldsymbol{\ell}})\big\}^2 \bigg]\\
    &= \sum_{j \in \mathcal{J}} \int_{\mathcal{X}_j}  \big\{\hat{f}^{\text{post}}_{j}(x) - \hat{f}^{\text{pre}}_{j}(x)\big\}^2 \mathrm{dF}_{X_j}(x) + 
    \sum_{\boldsymbol{\ell} \in \mathcal{L}} \int_{\mathcal{X}_{\boldsymbol{\ell}}} \big\{\hat{f}^{\text{post}}_{\boldsymbol{\ell}}(\mathbf{x}) - \hat{f}^{\text{pre}}_{\boldsymbol{\ell}}(\mathbf{x})\big\}^2 \mathrm{dF}_{\mathbf{X}_{\boldsymbol{\ell}}}(\mathbf{x}).
\end{align*}
We choose the squared distance here because large deviations between shape functions must be penalised; other options, such as the absolute distance, may also be used. Although the ESS could adequately measure structural stability in theory, it is computationally intractable, especially for large~$p$. We thus resort to its empirical version.

We introduce the Mean Squared Shape function distance (MSS), a data-dependent structural distance metric for functionally transparent models. The MSS is designed to provide a numerically efficient empirical approximation of the ESS that supports model-, observation-, and feature-level stability assessments without requiring assumptions on the covariate distributions. Formally, in a dataset $\mathcal{D}$, the individual MSS of observation $i \in \mathcal{I}$ (the contribution of observation~$i$ to the MSS) is
\begin{align}
    \text{MSS}_i
    &= \sum_{j \in \mathcal{J}}\frac{1}{|\mathcal{D}|}\big\{\hat{f}^{\text{post}}_{j}(x_{ij}) - \hat{f}^{\text{pre}}_{j}(x_{ij})\big\}^2 + \sum_{\boldsymbol{\ell} \in \mathcal{L}}\frac{1}{|\mathcal{D}|} \big\{\hat{f}^{\text{post}}_{\boldsymbol{\ell}}(\mathbf{x}_{i\boldsymbol{\ell}}) - \hat{f}^{\text{pre}}_{\boldsymbol{\ell}}(\mathbf{x}_{i\boldsymbol{\ell}})\big\}^2 \label{eq:mss_i} \\
    &= \sum_{j \in \mathcal{J}} \text{MSS}_{ij}  + 
    \sum_{\boldsymbol{\ell} \in \mathcal{L}} \text{MSS}_{i\boldsymbol{\ell}},
    \notag
\end{align}
where $\text{MSS}_{ij}$ and $\text{MSS}_{i\boldsymbol{\ell}}$ are respectively the contributions of main effect $j \in \mathcal{J}$ and interaction~$\boldsymbol{\ell} \in \mathcal{L}$ to~$\text{MSS}_{i}$. The $j$th main effect MSS (the contribution of main effect $j$ to the MSS), obtained by summing its per-observation contributions, is
\begin{align}
    &\text{MSS}_{j}^{\text{main}} =  \sum_{i \in \mathcal{I}} \text{MSS}_{ij}.
    \label{eq:mss_i_bias_j}
\end{align}
Similarly, the interaction MSS for pair $\boldsymbol{\ell}$ (the contribution of interaction $\boldsymbol{\ell}$ to the MSS) is
\begin{align}
    &\text{MSS}_{\boldsymbol{\ell}}^{\text{int}} =  \sum_{i \in \mathcal{I}} \text{MSS}_{i\boldsymbol{\ell}}.
    \label{eq:mss_i_bias_l}
\end{align}
For simplicity, MSS for all main effects and interactions together are referred to as per-feature contributions to the MSS. From Equations~\eqref{eq:mss_i}~to~\eqref{eq:mss_i_bias_l},
we see that the MSS admits an additive decomposition of the model instability at the observation and feature levels:
\begin{align}
\label{eq:decomposition_MSS}
    \text{MSS}  &= \sum_{i \in \mathcal{I}} \text{MSS}_i = 
    \sum_{j \in \mathcal{J}} \text{MSS}_{j}^{\text{main}} +  \sum_{\boldsymbol{\ell} \in \mathcal{L}} \text{MSS}^{\text{int}}_{\boldsymbol{\ell}}.
\end{align}
For brevity, Equations~\eqref{eq:mss_i}~to~\eqref{eq:decomposition_MSS} omit their common arguments $\big(\hat{\mathcal{F}}^{\text{pre}}, \hat{\mathcal{F}}^{\text{post}}; \mathcal{D}\big)$. To express the MSS on the scale of the log prediction (when using a log link), we compute the Root Mean Squared Shape function distance (RMSS), defined as the square root of the MSS.

\begin{remark}
Given that some functionally transparent models handle categorical features by one-hot encoding, we must account for changes in reference levels during retraining. In GLMs or GAMs, coefficients related to categorical features  represent the expected change in the outcome associated with that category, using the reference level as the baseline. Inconsistent baselines across updates would result in incomparable pre- and post-update shape functions, making the MSS no longer meaningful. Therefore, before computing our metric, we enforce identical pre- and post-update reference levels; predictive stability and performance are unaffected.
\end{remark}

\subsection{Predictive stability and performance metrics}
\label{sec:existing_metrics}

We now present the other metrics monitored in the case studies. To assess predictive stability, we use the mean relative dislocation (MRD), a metric inspired by industry practice. It is defined as follows
\begin{align*}
    &\text{MRD} \big(\hat{\mathcal{F}}^{\text{pre}}, \hat{\mathcal{F}}^{\text{post}}; \mathcal{D}\big) = \frac{1}{|\mathcal{D}|} \sum_{i \in \mathcal{I}} \bigg|\frac{\hat{\mathcal{F}}^{\text{post}}(\mathbf{x}_i) - \hat{\mathcal{F}}^{\text{pre}}(\mathbf{x}_i)}{\hat{\mathcal{F}}^{\text{pre}}(\mathbf{x}_i)} \bigg|,
\end{align*}
where the $i$th term in the sum is the relative dislocation of observation $i$. These relative dislocations enable stability assessment at the observation level. Predictive stability metrics, such as the MRD, measure the distance between predictions from two different models. In \appref{app:divide_and_conquer}, we derive lower and upper bounds for the absolute dislocation (the absolute difference between the pre- and post-update predictions) of functionally transparent models.

We also monitor predictive performance using the deviance, a common metric to measure the distance between the post-update predictions and the true responses. It accounts for the distributional assumptions made on the response and thus handles asymmetry. For example, the Poisson deviance is used for count data, the gamma deviance for strictly positive right-skewed data, and the Tweedie deviance for semi-continuous data with a point mass at zero. Using $v \in (1,2)$ as the Tweedie power parameter, these deviances are defined as follows
\begin{align*}
    &\text{Dev}_{\text{Poisson}}^{\text{post}} \big(\hat{\mathcal{F}}^{\text{post}} ; \mathcal{D}\big) = \frac{2}{|\mathcal{D}|} \sum_{i \in \mathcal{I}}  \Bigg[ y_i \ln\Bigg\{\frac{y_i}{\hat{\mathcal{F}}^{\text{post}}(\mathbf{x}_i)}\Bigg\} - y_i + \hat{\mathcal{F}}^{\text{post}}(\mathbf{x}_i) \Bigg],\\
    &\text{Dev}_{\text{gamma}}^{\text{post}} \big(\hat{\mathcal{F}}^{\text{post}}; \mathcal{D} \big) = \frac{2}{|\mathcal{D}|} \sum_{i \in \mathcal{I}}  \Bigg[ - \ln\Bigg\{\frac{y_i}{\hat{\mathcal{F}}^{\text{post}}(\mathbf{x}_i)}\Bigg\} + \frac{y_i - \hat{\mathcal{F}}^{\text{post}}(\mathbf{x}_i)}{\hat{\mathcal{F}}^{\text{post}}(\mathbf{x}_i)} \Bigg],\\
    &\text{Dev}_{\text{Tweedie}}^{\text{post}} \big(\hat{\mathcal{F}}^{\text{post}};\mathcal{D} \big) = \frac{2}{|\mathcal{D}|} \sum_{i \in \mathcal{I}}  \Bigg[ y_i \Bigg\{ \frac{y_i^{1-v} - \hat{\mathcal{F}}^{\text{post}}(\mathbf{x}_i)^{1-v}}{1-v}\Bigg\} - \Bigg\{ \frac{y_i^{2-v} - \hat{\mathcal{F}}^{\text{post}}(\mathbf{x}_i)^{2-v}}{2-v}\Bigg\} \Bigg].
\end{align*}
For further details on the use of deviance in machine learning, see, for example, \citet{denuit2019effective, henckaerts2021boosting, wuthrich2023statistical}.

\section{Stabilisation strategies}
\label{sec:control}

The MSS can be used to assess structural stability across model classes, much in the same way deviance is used to assess predictive performance. Comparing vanilla 
algorithms yields a ranking of model classes and assists the selection of a suitable one for a specific problem. However, stability can also vary within a given model class depending on modelling choices. Several heuristic methods, which we refer to as stabilisation strategies, are available to control stability within model classes, and our metric enables a quantitative assessment of their impact. In this section, we briefly present the two stabilisation strategies considered in our case studies: sequential modelling, and fixed hyperparameters and/or interaction terms during the retraining.

\begin{figure}[t!]
    \centering
    \makebox[\textwidth][c]{%
      \includegraphics[width = \textwidth]{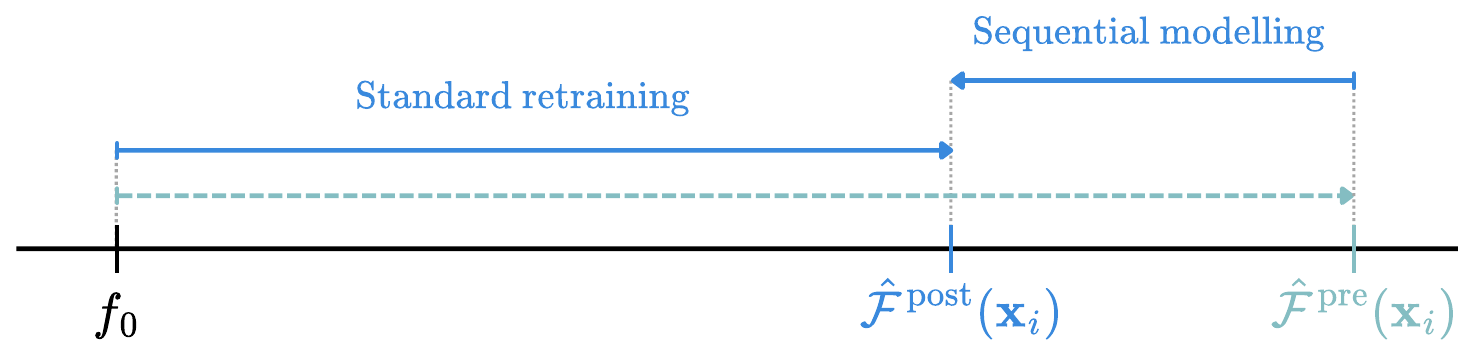}}
    \caption{Simplified comparison of standard retraining and sequential modelling.}
    \label{fig:viz_seq_mod}
\end{figure}

A popular stabilisation strategy consists of offsetting the post-update model by the predictions of the pre-update model, which is equivalent to using the pre-update residuals as the response variable in the model retraining. This approach, referred to as sequential modelling in \citet{yan2009applications}, is also viewed as adding a model layer on top of an existing rating plan. For instance, common main effects may be imposed across regions through a preliminary model, while a subsequent model, built on top, captures residual effects specific to each region. An additional motivation is to model main effects separately from interaction effects, thereby facilitating estimation. For our case studies, sequential modelling anchors the post-update model around the pre-update predictions, as illustrated in Figure~\ref{fig:viz_seq_mod}. Arrows represent the learning processes, from starting value to fitted prediction. Specifically, the dashed arrow represents the learning process of the pre-update model; it is stand-alone. The solid arrows represent two options for the learning process of the post-update model; the starting values differ depending on whether sequential modelling is used or not. Under standard retraining, the learning process of the post-update model starts at $f_0$, the prediction at initialisation, like the pre-update model. With sequential modelling, it instead starts at the pre-update model’s predictions (the anchor). The rationale is that this anchoring should lead to smaller updates of the shape functions.

\newpage

Another strategy is to fix hyperparameters and/or selected interaction terms. During model retraining, hyperparameters are generally selected anew based on the post-update data. For a transparent model such as EBM, whose structure is highly sensitive to tuning, selecting different hyperparameters may reduce both predictive and structural stability. Any gain in predictive performance should be large enough to justify the computational cost of hyperparameter tuning.

Among the three transparent models that we will consider (GLM, GAM, and EBM), only EBM includes interaction terms by default. Because the selected interactions may differ between the pre-update and post-update models, they constitute an additional source of structural instability. Fixing the set of selected interaction terms across updates, while still re-estimating the associated shape functions during retraining, may improve stability.

\section{Auto insurance case studies}
\label{sec:case_study}

The main goal of this section is to illustrate the usefulness of our metric. On three public datasets and four model classes, we emulate an update using Algorithm~\ref{alg:data-split-variation} ($\alpha=0.7, \beta=0.3$) in order to monitor the stability and performance metrics from Section~\ref{sec:stability_metrics}.  We investigate the impact of two stabilisation strategies on these metrics, and find that EBM is a functionally transparent model that combines stability with strong predictive performance.

We present the selected datasets and model classes in Section~\ref{sec:datasets}. For each combination of model class, dataset, and response, we study the structural stability, the predictive stability, and the predictive performance in Section~\ref{sec:results}. We assess, in Section~\ref{sec:impact_control}, the impact of two stabilisation strategies on the metrics considered. We provide details about the training and retraining procedures in \appref{app:train_retrain_metho}.

\subsection{Datasets and model classes}
\label{sec:datasets}
 
Three publicly available auto insurance datasets from the \textsf{R} package \texttt{CASdatasets} \citep{dutang2024casdatasets} are considered: \texttt{norauto}, \texttt{beMTPL97}, and \texttt{pg15training}. In Table~\ref{tab:datasets}, we detail their number of policyholders $n$, features retained\footnote{Two variables with more than 450 levels were removed prior to this analysis.} $p$, and categorical features $p_{\text{cat}}$. All datasets also comprise an exposure-to-risk, which corresponds to the duration, in days or years, for which the policyholder held insurance coverage. 

\begin{table}[t!]
    \centering
        \caption{Overview of the datasets.}
    \begin{tabular}{lrrr}
        \toprule
        Dataset & $n$ & $p$ & $p_{\text{cat}}$ \\
        \midrule
        \texttt{norauto} & 183~999 & 4 & 4 \\[4pt]
        \texttt{beMTPL97} & 163~212 & 11 & 5 \\[4pt]
        \texttt{pg15training} & 100~021 & 13 & 7 \\
        \bottomrule
    \end{tabular}
    \label{tab:datasets}
\end{table}

For each dataset, we model the yearly claim frequency, the claim severity (the average claim cost given that at least one claim was reported), and the loss cost (the total paid amount over the exposure period). This gives rise to three regression problems for which we assume a log link along with Poisson, gamma, and Tweedie distributions, respectively. The varying exposure-to-risk are considered with offsets in the Poisson and Tweedie regressions.

We select the functionally transparent model classes: GLM, fitted with base \textsf{R}, GAM, implemented in \textsf{R} through \texttt{mgcv} \citep{wood2003thin}, and EBM with its \texttt{Python} distribution in \texttt{interpret}
\citep{nori2019interpretmlunifiedframeworkmachine}. The GLM is the traditional benchmark in actuarial practice. In both the training and test sets, we center the numerical covariates by subtracting their training sample mean. This preprocessing enforces centred shape functions, ensuring consistency of GLM with GAM and EBM, where such centering arises naturally from identifiability constraints. It also induces modularity, meaning that shape functions can be removed from the prediction function (i.e., set to zero) without introducing bias in the predictions \citep{caruana2015intelligible}. The GAM, which increases flexibility over the GLM by fitting thin plate regression splines on the numerical features, serves as a bridge between GLM and EBM. The latter is the state-of-the-art transparent model. We finally include XGBoost in its \textsf{R} implementation \texttt{xgboost} \citep{chen2019package} to benchmark predictive stability and performance. However, tree depths greater than two in XGBoost, introducing high-order interactions, preclude the quantification of its structural stability with the MSS. 

\subsection{Multilevel stability analysis}
\label{sec:results}

We now explain that our structural stability metric enables a deeper analysis than predictive stability metrics. We start with the model-level and observation-level results in Sections~\ref{sec:model_level} and~\ref{sec:observation_level}; both structural and predictive stability metrics are studied at these depths. In Section~\ref{sec:feature_level}, however, we provide a feature-level analysis, which is only feasible with exactitude using our structural stability metric on functionally transparent models. We focus on \texttt{beMTPL97} given the consistency of findings across datasets, but results for all datasets are provided in \appref{app:model_level_results}.

\subsubsection{Model-level analysis}
\label{sec:model_level}

To compare model classes, we report the average rank of each metric over the three responses (claim frequency, claim severity, and loss cost) in Table \ref{tab:avg_ranks_beMTPL97}. These are computed from the metrics on the post-update test datasets given in Table~\ref{tab:combined_all} of \appref{app:model_level_results}. Recall that XGBoost is not functionally transparent, so its RMSS cannot be computed. As seen in Table~\ref{tab:avg_ranks_beMTPL97}, GLM is the most structurally stable across all responses, whereas GAM is the least. In terms of predictive stability, GLM again ranks first, while XGBoost ranks last on average; more flexible models tend to have lower predictive stability and, unsurprisingly, EBM and XGBoost achieve the best average performance.

\begin{table}[b!]
\centering
\caption{Average rank of each metric over the three responses on \texttt{beMTPL97}.}
\begin{tabular}{lccc}
\toprule
Model 
& RMSS & MRD & Dev \\
\midrule
GLM      & \textbf{1.0}          & \textbf{1.0} & 3.3          \\
GAM      & 3.0          & 2.7          & 2.7          \\
EBM      & 2.0 & 3.0          & \textbf{2.0} \\
XGBoost  & $-$          & 3.3          & \textbf{2.0} \\
\bottomrule
\end{tabular}
\label{tab:avg_ranks_beMTPL97}
\end{table}

\begin{figure}[t!]
    \centering
    \makebox[\textwidth][c]{%
      \includegraphics[width = \textwidth]{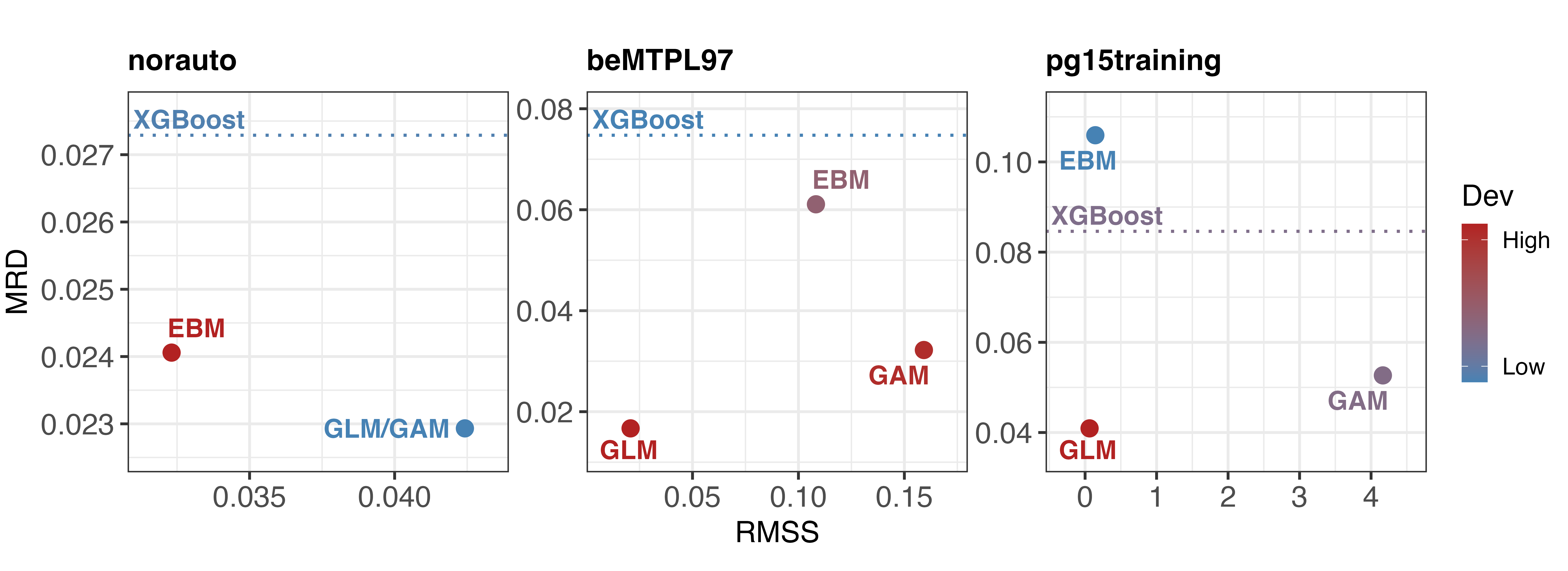}}
    \caption{Relation between the MRD and the RMSS coloured by Poisson deviance for each dataset.}
    \label{fig:rmss_versus_mrd_poisson}
\end{figure}

In Figure~\ref{fig:rmss_versus_mrd_poisson}, we show the MRD in terms of the RMSS, and the Poisson deviance as the relative colour scale for all frequency models on each dataset. The GLM is the most stable model in almost every case: it has the lowest MRD on all three datasets and the lowest RMSS on \texttt{beMTPL97} and \texttt{pg15training}. The exception is \texttt{norauto}, where the EBM is more structurally stable. No single model wins on predictive performance. The GLM is on par with XGBoost on \texttt{norauto}, XGBoost is best on \texttt{beMTPL97}, and the EBM is best on \texttt{pg15training}. Complete results are in Table~\ref{tab:avg_ranks_all} of \appref{app:model_level_results}. Broadly speaking, flexibility improves performance at the expense of stability. Our case studies suggest that EBM offers a compromise between predictive power and stability. 

\subsubsection{Observation-level analysis}
\label{sec:observation_level}

To deepen the analysis, we use both structural and predictive stability metrics to decompose model instability into per-observation contributions. Specifically, we split the MSS into individual MSS and the MRD into relative dislocations. We focus on the frequency in the \texttt{beMTPL97} dataset.

\begin{figure}[t!]
    \centering
    \makebox[\textwidth][c]{%
      \includegraphics[width = \textwidth]{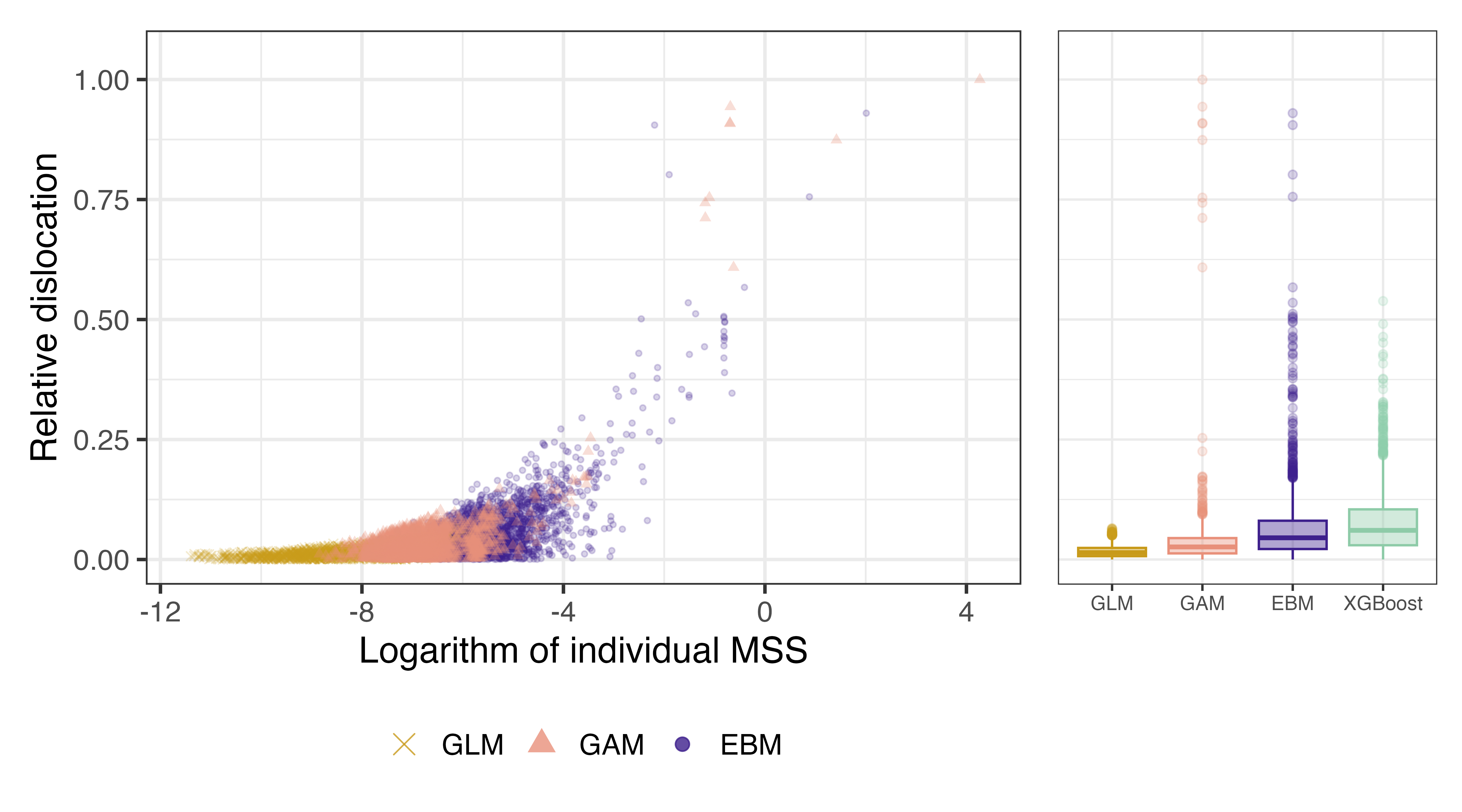}}
        \caption{
        Relative dislocation in terms of the logarithm of individual MSS for the GLM (cross), GAM (triangle), and EBM (circle) frequency models on \texttt{beMTPL97} (left). Corresponding boxplot of relative dislocation by model class (right). Both panels are based on the same subset of observations from the post-update test dataset.}
    \label{fig:logmssi_versus_rdi_poisson}
\end{figure}

In the left panel of Figure~\ref{fig:logmssi_versus_rdi_poisson}, we report the relative dislocations in terms of the log-transformed individual MSS for a random subset of test observations. Compared to the other transparent models, the GLM exhibits lower individual MSS and relative dislocation for most observations, as expected from its greater stability.
The observation-level metrics suggest that a low individual MSS is associated with a low relative
dislocation, whereas a high individual MSS does not necessarily imply a high relative dislocation. If
the structure remains unchanged (low individual MSS), the predictions must also remain unchanged
(low relative dislocation). However, changes in the structure do not necessarily transfer to the
predictions, because the differences in shape functions across features may offset each other. Some severely unstable observations make the MSS of the GAM higher than that of EBM. To understand this behaviour of the GAM, we push the analysis to the feature level in Section~\ref{sec:feature_level}.

In the right panel of Figure~\ref{fig:logmssi_versus_rdi_poisson}, we show the boxplot of relative dislocations for all models to compare their predictive stability. The interquartile range increases with
model flexibility. The especially low volatility of the GLM’s relative dislocations reflects its high predictive
stability. The quartiles of XGBoost, the black box model, are higher than their
counterpart for EBM, the state-of-the-art transparent model; high flexibility is associated with low
predictive stability. For all models, we observe several outliers, suggesting heavy right tails in the
distributions of relative dislocations.

\newpage
\subsubsection{Feature-level analysis}
\label{sec:feature_level}

Decomposing model- and observation-level instability into per-feature contributions enables a more granular assessment of stability. However, only the MSS allows such decomposition with exactitude---for black box algorithms, we must resort to post-hoc explanation stability \citep{zhao2026explainable}, which is only approximate. Below, we revisit our previous analysis of the claim frequency models on \texttt{beMTPL97} by focusing on feature-wise contributions to the structural instability.

\begin{figure}[b!]
    \centering
    \makebox[\textwidth][c]{%
      \includegraphics[width = 0.95\textwidth]{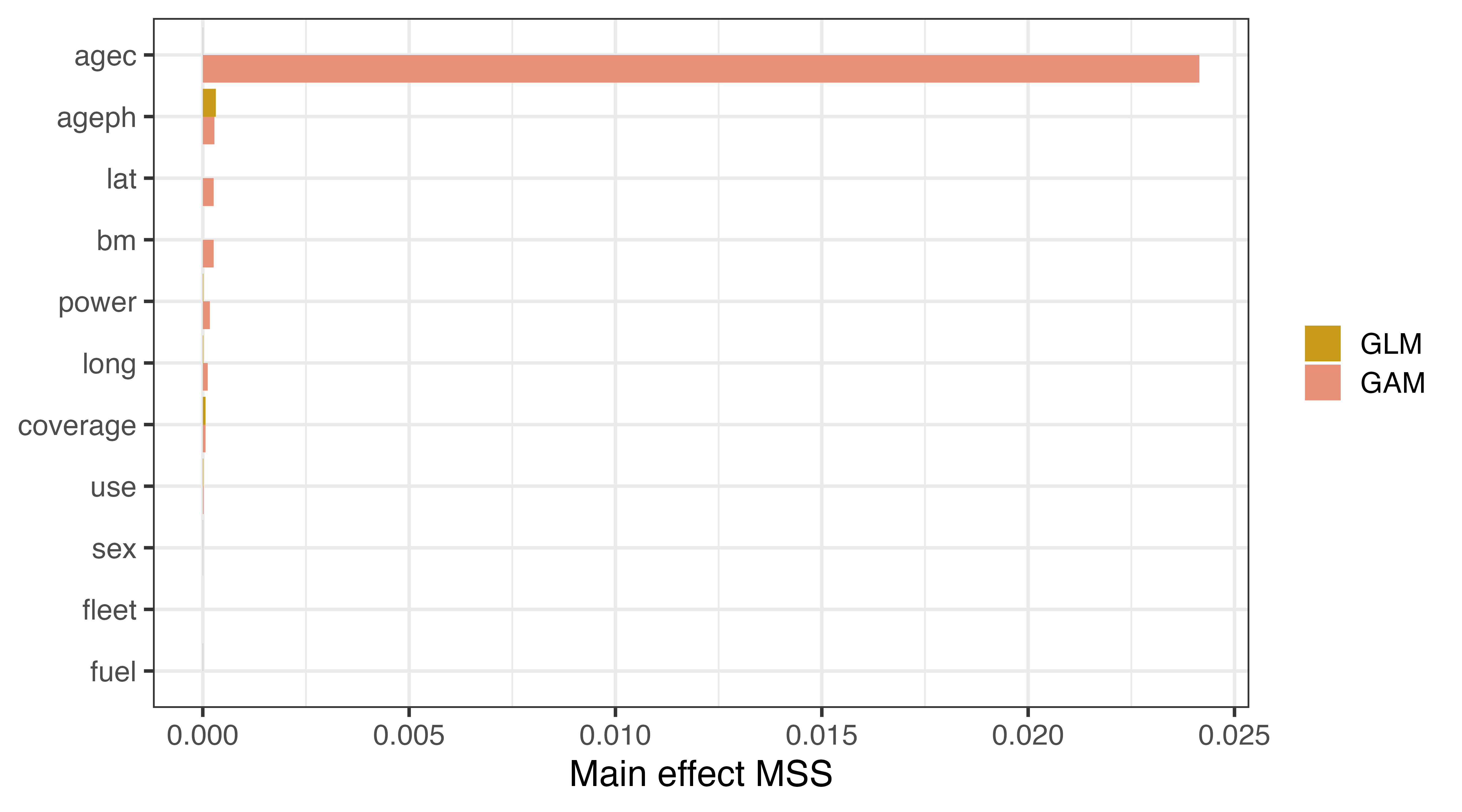}}
    \caption{Main effect MSS for frequency GLM and GAM on \texttt{beMTPL97}.}
    \label{fig:glm_gam_by_feat_model}
\end{figure}

\begin{figure}[b!]
    \centering
    \makebox[\textwidth][c]{%
      \includegraphics[width = 0.95\textwidth]{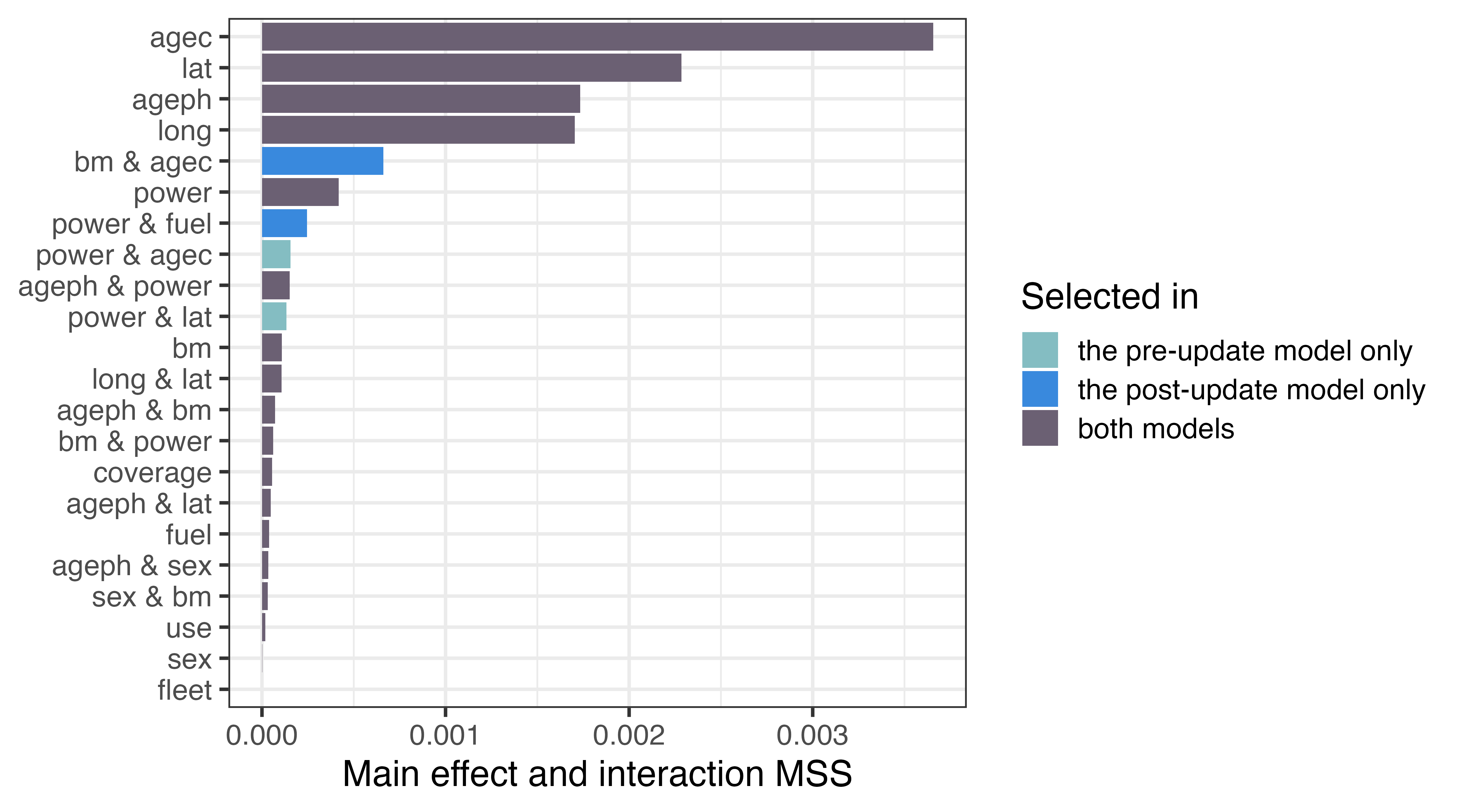}}
    \caption{Main effect and interaction MSS for frequency EBM on \texttt{beMTPL97}.}
    \label{fig:ebm_by_feat_model}
\end{figure}

In Figure~\ref{fig:glm_gam_by_feat_model}, we show the main effect MSS for the GLM and GAM. All main effects are included in both models; no feature selection is performed by default. The MSS of \texttt{agec} is by far the largest contributor to the instability of the GAM, while the GLM exhibits low instability across all features. This contrasts with the results in Figure~\ref{fig:ebm_by_feat_model}, where many main effects and interactions contribute to the MSS for EBM. By design, all main effects are included in EBM; among these, four out of six numerical main effects (\texttt{agec}, \texttt{lat}, \texttt{ageph}, and \texttt{long}) account for most of the instability. As highlighted by the colours in Figure~\ref{fig:ebm_by_feat_model}, most interactions with high contribution to the MSS are selected in only one of the pre- or post-update models. This reflects the structural stability cost of inconsistent variable or interaction selection across updates.

\begin{figure}[t!]
    \centering
    \makebox[\textwidth][c]{%
      \includegraphics[width = 0.945\textwidth]{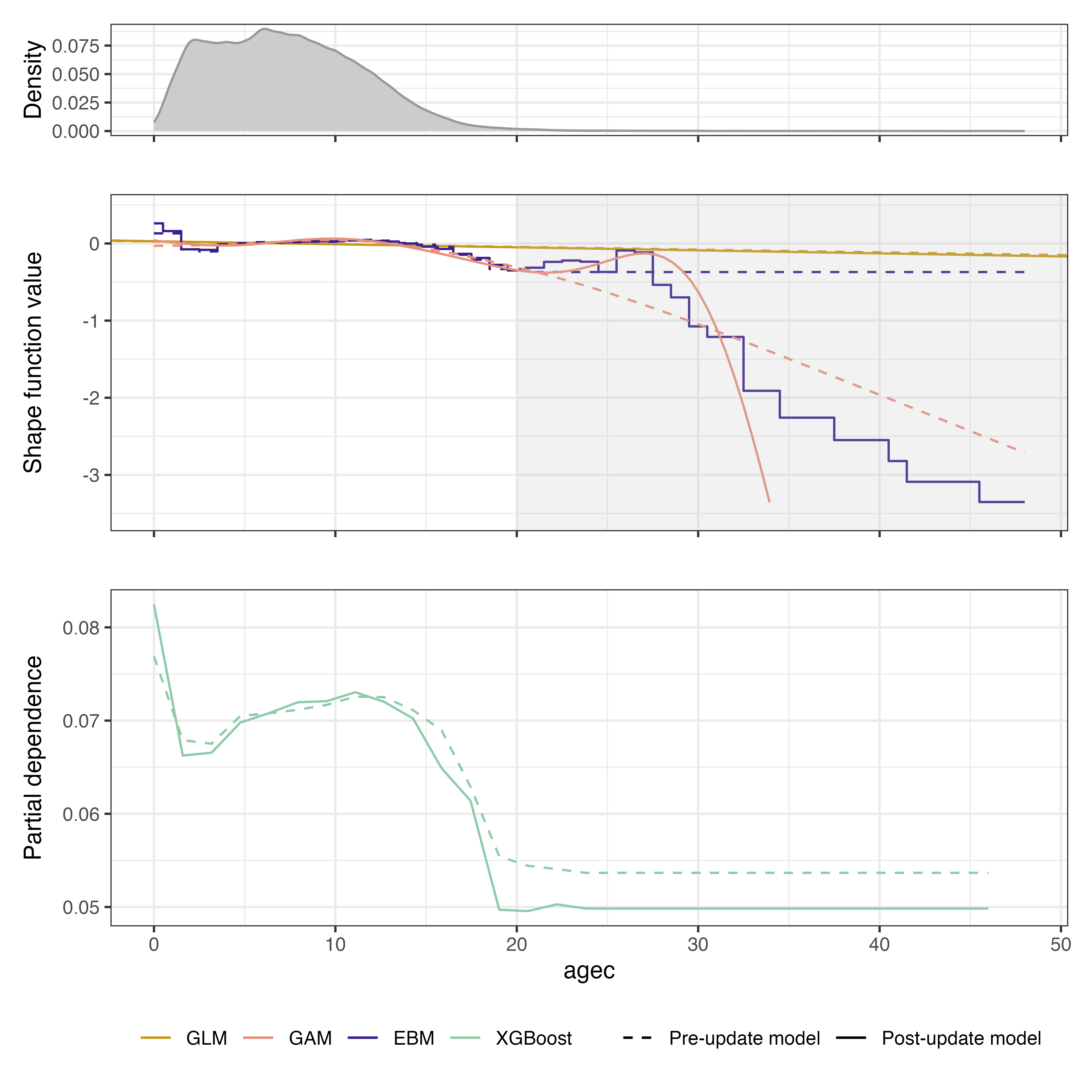}}
    \caption{Feature density (top), pre- and post-update shape functions from the three transparent models (middle), and pre- and post-update partial dependence plots from XGBoost (bottom) for \texttt{agec} with the frequency in \texttt{beMTPL97}.}
    \label{fig:shape_functions_agec}
\end{figure}

The feature-level MSS provides guidance when investigating the sources of structural instability. Our analysis of Figures~\ref{fig:glm_gam_by_feat_model}~and~\ref{fig:ebm_by_feat_model} suggests that \texttt{agec}, \texttt{lat}, and \texttt{ageph} deserve further study. The first two are examined below and the last is addressed in \appref{app:instability-driving-features}. 

We analyse \texttt{agec} in Figure~\ref{fig:shape_functions_agec}. Specifically, we display the feature density on the post-update training dataset (top), the pre- and post-update shape functions from the three transparent models (middle), and the pre- and post-update partial dependence plots from XGBoost (bottom). In the middle panel, we see that the GLM shape functions are similar across the domain, reflecting its negligible MSS for \texttt{agec}. For GAM and EBM, the instability arises in the right tail of the vehicle age distribution (see shaded region). The bottom panel serves to compare the trend of the XGBoost partial dependence functions with those of the shape functions in the middle panel. However, the distance between the dashed and solid lines are not on the same scale in the middle and bottom panels. Still, the trends suggest an agreement on the marginal effect of \texttt{agec} across models. We stress that the explicit visualisation of the shape functions is a particularity of functionally transparent models.

\begin{figure}[t!]
    \centering
    \makebox[\textwidth][c]{%
      \includegraphics[width = 0.945\textwidth]{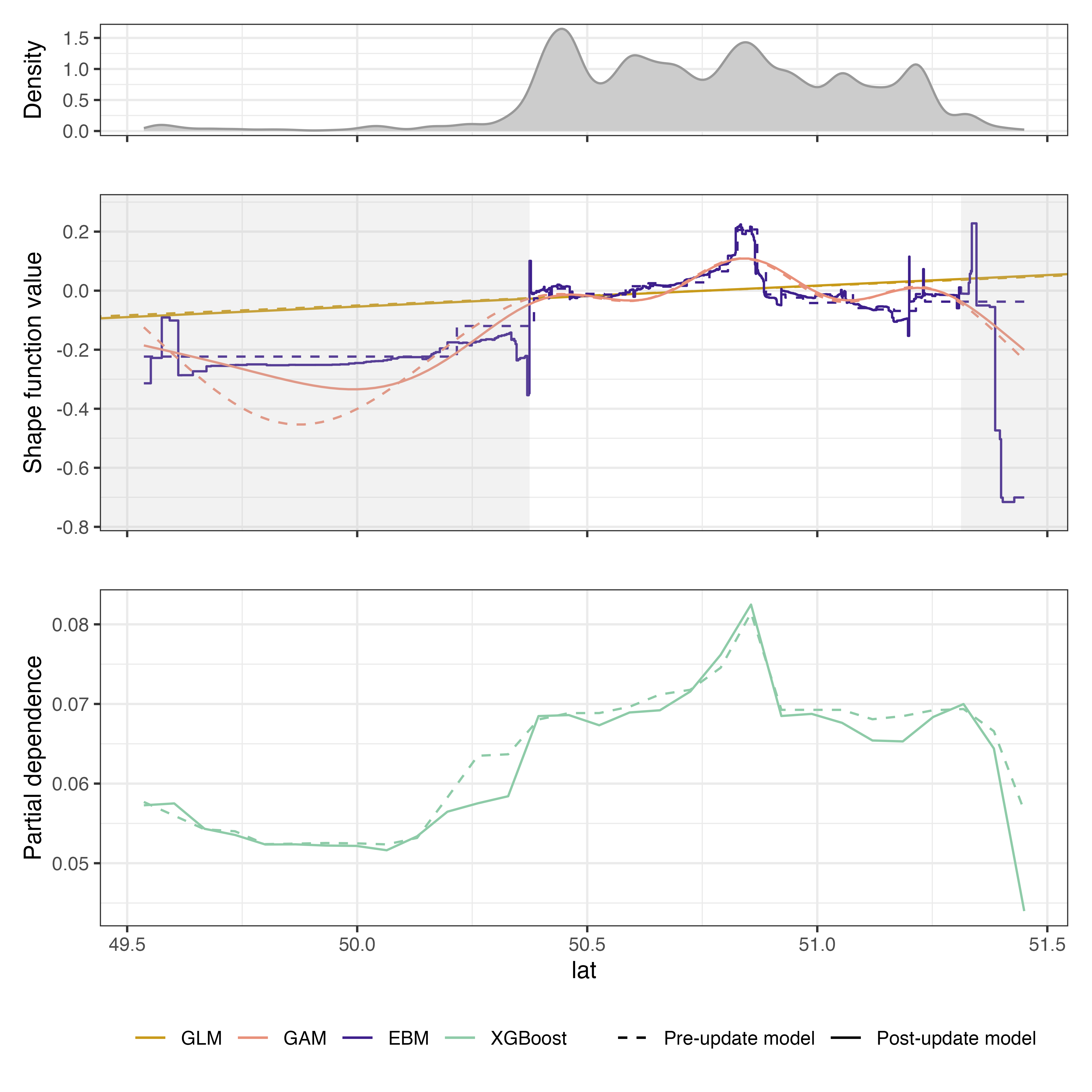}}
    \caption{Feature density (top), pre- and post-update shape functions from the three transparent models (middle), and pre- and post-update partial dependence plots from XGBoost (bottom) for \texttt{lat} with the frequency in \texttt{beMTPL97}.}
    \label{fig:shape_functions_lat}
\end{figure}

The second-highest contributor to the EBM instability, \texttt{lat}, is analysed in a similar format in Figure~\ref{fig:shape_functions_lat}. For this feature, the instability of the GAM (smooth curves) and EBM (step functions) primarily originates from extrapolation; the pre- and post-update shape functions of both models differ most in the shaded regions, where there are fewer observations. The GLM shape functions are superimposed, yielding a near zero MSS for \texttt{lat}. According to its post-hoc explanation, XGBoost generally agrees with the functionally transparent models on the marginal effect of this covariate.

\subsection{Impact of stabilisation strategies}
\label{sec:impact_control}

So far, we have only considered the vanilla models. We now examine the impact of different stabilisation strategies on the stability and performance of GLM, EBM, and XGBoost for all datasets. The GAM is not considered further, because it never achieved the best average rank across any metric on any dataset. We study sequential modelling, and fixed hyperparameters and/or interaction terms during retraining. Detailed results on all datasets and all responses are provided in Tables~\ref{tab:seq_avg_ranks_all}--\ref{tab:ebm_same_int_rmss_mrd_deviance} in \appref{app:add_res_stab_strat}. Here, we present variations in the metrics induced by each stabilisation strategy, expressed as relative changes with respect to the associated baseline. A negative variation indicates an improvement in the metric due to the stabilisation strategy. Any discrepancy from calculations based on the detailed results in the appendix comes from rounding of the latter.

First, we focus on the results of sequential modelling on EBM and XGBoost; a discussion on the specificities of the GLM is provided in \appref{app:glm_seq_mod}. In Table~\ref{tab:ebm_xgb_without_with_seqmod}, we report the variations for all metrics when using sequential modelling. For both EBM and XGBoost, the MRD is reduced by at least 30\%, and up to 88\% in the best case. Sequential modelling halves the structural instability of EBM in eight of the nine dataset--response combinations; severity modelling on \texttt{norauto} is the sole exception with an increase in RMSS below five percent. Despite this substantial stability improvement, performance is essentially preserved, with reductions in deviance below one percent in most cases.

Second, for EBM and XGBoost, we retrain the post-update models using the hyperparameters of their corresponding pre-update model (see \appref{app:train_retrain_metho}). In Table~\ref{tab:ebm_xgb_fixedhyp_rmss_mrd_deviance}, we report metric variations when fixing hyperparameters. This strategy improves both predictive and structural stability in most cases for EBM, with improvements ranging from small to substantial. The associated cost is minimal, with predictive performance decreasing by often less than one percent. For XGBoost, predictive stability increases in seven of the nine cases, while performance remains essentially unchanged.

\begin{table}[b!]
\centering
\footnotesize
\caption{Variations (in \%) of each metric when using sequential modelling.}
\label{tab:ebm_xgb_without_with_seqmod}
{\setlength{\tabcolsep}{2pt}
\begin{tabular}{llrrrrrrrrr}
\toprule
& & \multicolumn{3}{c}{\texttt{norauto}} & \multicolumn{3}{c}{\texttt{beMTPL97}} & \multicolumn{3}{c}{\texttt{pg15training}} \\
\cmidrule(lr){3-5} \cmidrule(lr){6-8} \cmidrule(lr){9-11}
Model & Metric & Frequency & Severity & Loss cost & Frequency & Severity & Loss cost & Frequency & Severity & Loss cost \\
\midrule
\multirow{1.2}{*}{EBM}
  & RMSS & $-$68.8 & $+$3.5 & $-$54.4 & $-$52.2 & $-$70.6 & $-$59.3 & $-$56.2 & $-$61.4 & $-$49.5 \\
  & MRD & $-$72.4 & $-$66.8 & $-$48.3 & $-$60.6 & $-$64.6 & $-$63.2 & $-$81.2 & $-$70.2 & $-$88.4 \\
  & Dev & $+$0.054 & $+$0.255 & $+$0.060 & $+$0.099 & $+$0.791 & $+$2.145 & $+$0.681 & $+$0.071 & $+$0.765 \\
\midrule
\multirow{1.2}{*}{XGBoost}
  & MRD & $-$31.5 & $-$59.3 & $-$31.1 & $-$72.5 & $-$50.3 & $-$60.2 & $-$75.9 & $-$36.3 & $-$73.6 \\
  & Dev & $+$0.007 & $+$0.406 & $+$0.050 & $-$0.030 & $+$0.620 & $+$0.259 & $+$0.308 & $-$0.128 & $+$0.209 \\
\bottomrule
\end{tabular}
}
\end{table}

\begin{table}[b!]
\centering
\footnotesize
\caption{Variations (in \%) of each metric when fixing hyperparameters.}
\label{tab:ebm_xgb_fixedhyp_rmss_mrd_deviance}
{\setlength{\tabcolsep}{2pt}
\begin{tabular}{llrrrrrrrrr}
\toprule
& & \multicolumn{3}{c}{\texttt{norauto}} & \multicolumn{3}{c}{\texttt{beMTPL97}} & \multicolumn{3}{c}{\texttt{pg15training}} \\
\cmidrule(lr){3-5} \cmidrule(lr){6-8} \cmidrule(lr){9-11}
Model & Metric & Frequency & Severity & Loss cost & Frequency & Severity & Loss cost & Frequency & Severity & Loss cost \\
\midrule
\multirow{1.2}{*}{EBM}
  & RMSS & $-$9.1   & $+$2.6   & $-$0.4   & $-$33.5  & $-$10.6  & $-$41.4  & $-$8.6   & $-$28.0  & $-$3.9   \\
  & MRD  & $-$4.6   & $-$2.9  & $-$0.9   & $-$14.7  & $-$6.5   & $-$37.6  & $-$15.7  & $-$20.5  & $-$3.4   \\
  & Dev  & $+$0.006 & $+$0.023 & $+$0.014 & $+$0.097 & $+$0.053 & $+$1.136 & $+$0.246 & $+$0.012 & $+$0.009 \\
\midrule
\multirow{1.2}{*}{XGBoost}
  & MRD  & $-$1.7   & $+$0.2   & $-$4.9   & $-$46.4  & $-$0.4   & $-$9.6   & $-$2.6   & $-$1.2   & $+$0.3   \\
  & Dev  & $+$0.001 & $+$0.021 & $-$0.003 & $-$0.021 & $+$0.091 & $-$0.025 & $+$0.143 & $-$0.057 & $+$0.284 \\
\bottomrule
\end{tabular}}
\end{table}

\begin{table}[b!]
\footnotesize
\centering
\caption{Variations (in \%) of each metric for EBMs when fixing hyperparameters and interaction terms.}
\label{tab:ebmfixedhyp_same_int_rmss_mrd_deviance}
{\setlength{\tabcolsep}{2pt}
\begin{tabular}{lrrrrrrrrr}
\toprule
& \multicolumn{3}{c}{\texttt{norauto}} & \multicolumn{3}{c}{\texttt{beMTPL97}} & \multicolumn{3}{c}{\texttt{pg15training}} \\
\cmidrule(lr){2-4} \cmidrule(lr){5-7} \cmidrule(lr){8-10}
& Frequency & Severity & Loss cost & Frequency & Severity & Loss cost & Frequency & Severity & Loss cost \\
\midrule
RMSS & $-$0.8   & $-$0.2   & $-$0.7   & $-$16.5  & $+$0.1   & $-$7.3   & $-$5.5   & $-$4.1   & $-$3.9   \\
MRD  & $+$0.1   & $+$2.5   & $-$0.9   & $-$17.1  & $+$0.1   & $-$8.5   & $-$7.4   & $-$2.7   & $-$5.1   \\
Dev  & $+$0.001 & $+$0.027 & $-$0.012 & $-$0.008 & $+$0.002 & $-$0.167 & $-$0.003 & $-$0.051 & $+$0.008 \\
\bottomrule
\end{tabular}}
\end{table}

Third, for EBM, we retrain the post-update models by fixing the hyperparameters and using the same set of interactions as selected in their corresponding pre-update model. In Table~\ref{tab:ebmfixedhyp_same_int_rmss_mrd_deviance}, we report metric variations with this strategy. The impact of fixing selected interactions is often negligible. However, for the frequency and the loss cost on \texttt{beMTPL97}, and for all responses on \texttt{pg15training}, maintaining the same interaction terms increases stability. As shown in Table~\ref{tab:ebm_same_int_rmss_mrd_deviance} of \appref{app:add_res_stab_strat}, this is true only when hyperparameters are also fixed; the impact of fixed interaction terms on stability is ambiguous when hyperparameters are allowed to vary.

\clearpage
\section{Conclusion}
\label{sec:conclusion}

In this paper, we introduce the notion of functional transparency. Then, we propose the Mean Squared Shape function distance, a data-dependent structural distance metric for functionally transparent models. Unlike existing predictive stability metrics, the MSS enables an exact feature-level assessment of stability, on top of the usual model- and observation-level assessments. To demonstrate the usefulness of our metric, we develop a framework to emulate model updates and conduct a benchmark study using three publicly available auto insurance datasets. We find that GLMs with centred numerical covariates are structurally and predictively stable, and that explainable boosting machines offer an interesting trade-off between stability and performance, while being functionally transparent. This is especially relevant for critical decision making and in highly-regulated industries, because the analyst can directly visualise, interpret and post-process the prediction functions. We also show that stabilisation strategies, such as sequential modelling and fixed hyperparameters, can substantially improve stability for both EBM and XGBoost models, with minimal impact on performance.

Our approach is not limited to general insurance and can be broadly applied, especially in critical decision applications, where transparency is often as important as performance. As described by \citet{breiman2001statistical}, the Rashomon effect, also known as the multiplicity of good models, states that several models can be equivalent in terms of predictive performance yet differ substantially in structure. If we can fully or partially identify the Rashomon set (i.e., the set of equally predictive models), insights provided by the MSS can then guide the selection of a model whose structure supports meaningful interpretations at deployment and across updates with additional data.

\paragraph{Acknowledgments}
We gratefully acknowledge an anonymous insurance company for insightful discussions, and especially one of its analysts for their generous help throughout this work. This research was funded by the Natural Sciences and Engineering Research Council of Canada (ALLRP-606959-25, RGPIN-2019-04190) and the Chaire d’actuariat of Université Laval. Félix Côté thanks Olivier Côté for artistic guidance, reading suggestions, and inspiration. The authors declare that generative artificial intelligence tools
were used occasionally, and solely, to suggest improvements to the
authors’ own code for some visualisations, or to assist with translations and reformulation.

\bibliographystyle{apalike}
\bibliography{ref}

\appendix

\section{Alternate version of Algorithm~1}
\label{app:algo2}

When no timestamp is available, we replace the temporal split of Algorithm~\ref{alg:data-split} with a random split of the complete dataset, as shown in Algorithm~\ref{alg:data-split-variation}.

\begin{algorithm}[H]
    \caption{Construction of the pre- and post-update training and test datasets without time stamp.}
    \label{alg:data-split-variation}
    \renewcommand{\algorithmicrequire}{\textbf{Input:}}
    \renewcommand{\algorithmicensure}{\textbf{Output:}}
    \renewcommand{\algorithmicreturn}{\textbf{Return}}
    
    \begin{algorithmic}[1]

    \REQUIRE Complete dataset $\mathcal{D}^{\text{post}}$, train proportion $\alpha \in (0,1)$, update proportion $\beta \in (0,1)$.
    
    \ENSURE Pre-update and post-update training and test datasets.

    \STATE Set $n^{\text{post}} := |\mathcal{D}^{\text{post}}|$.

    \STATE Sample the pre-update dataset $\mathcal{D}^{\text{pre}}$ by drawing $\lfloor (1 - \beta) n^{\text{post}}\rfloor$ observations from $\mathcal{D}^{\text{post}}$.

    \STATE Extract the update dataset $\mathcal{D}^{\text{up}} = \mathcal{D}^{\text{post}}\backslash \mathcal{D}^{\text{pre}}$. 

    \STATE Perform steps 3 to 8 of Algorithm~\ref{alg:data-split}.
    
    \RETURN $\mathcal{D}^{\text{pre}}_{\text{train}}$, $\mathcal{D}^{\text{post}}_{\text{train}}$, $\mathcal{D}^{\text{pre}}_{\text{test}}$, $\mathcal{D}^{\text{post}}_{\text{test}}$.
    
    \end{algorithmic}
\end{algorithm}

\newpage
\section{Divide and conquer}
\label{app:divide_and_conquer}

Sharp bounds for the absolute dislocation can be derived for functionally transparent models. We thereby establish a link between predictive and structural stability. Formally, assuming that the feature domains are independent, we define the Sharp Absolute Dislocation Upper bound (SADU) as
\begin{multline}
\text{SADU} \big(\hat{\mathcal{F}}^{\text{pre}}, \hat{\mathcal{F}}^{\text{post}}; \mathcal{X}\big) = g^{-1} \bigg[
\big|\hat{\beta}^{\text{post}}_{0} - \hat{\beta}^{\text{pre}}_{0}\big| + \\
\underset{\mathbf{x} \in \mathcal{X}}{\mathrm{sup}} \bigg\{ \sum_{j \in \mathcal{J}} \big|\hat{f}^{\text{post}}_{j}(x_j) - \hat{f}^{\text{pre}}_{j}(x_j)\big| + \sum_{\boldsymbol{\ell} \in \mathcal{L}} \big|\hat{f}^{\text{post}}_{\boldsymbol{\ell}}(\mathbf{x}_{\boldsymbol{\ell}}) 
    - \hat{f}^{\text{pre}}_{\boldsymbol{\ell}}(\mathbf{x}_{\boldsymbol{\ell}})\big| \bigg\}\bigg], \label{eq:optim_sharp_bound}
\end{multline}
where $\mathbf{x} = (x_j)_{j \in \mathcal{J}}$ and $\mathbf{x}_{\boldsymbol{\ell}} =  (x_{\ell})_{\ell \in \boldsymbol{\ell}}$. By changing the supremum for an infimum, we get the Sharp Absolute Dislocation Lower bound (SADL). Finding this sharp bound is a $p$-dimensional optimisation problem. It can be simplified through two approaches, both detailed in the remainder of this section. Throughout, we focus on the upper bound; the lower bound is obtained by replacing the supremum for an infimum in all subsequent equations.

The first approach computes loose bounds, reducing the problem to $p$ one-dimensional and $|\mathcal{L}|$ two-dimensional optimisations. By individually optimising over each term, main effect and interaction, we get the Loose Absolute Dislocation Upper bound (LADU), defined as
\begin{multline*}
    \text{LADU} \big(\hat{\mathcal{F}}^{\text{pre}}, \hat{\mathcal{F}}^{\text{post}}; \mathcal{X}\big) = g^{-1} \bigg[
    \big|\hat{\beta}^{\text{post}}_{ 0} - \hat{\beta}^{\text{pre}}_{0}\big| +\\
    \sum_{j \in \mathcal{J}} \underset{x_j \in \mathcal{X}_j}{\mathrm{sup}} \bigg\{\big|\hat{f}^{\text{post}}_{j}(x_j) - \hat{f}^{\text{pre}}_{j}(x_j)\big|\bigg\} + \sum_{\boldsymbol{\ell} \in \mathcal{L}} \underset{\mathbf{x}_{\boldsymbol{\ell}} \in \mathcal{X}_{\boldsymbol{\ell}}}{\mathrm{sup}} \bigg\{ \big|\hat{f}^{\text{post}}_{\boldsymbol{\ell}}(\mathbf{x}_{\boldsymbol{\ell}}) 
    - \hat{f}^{\text{pre}}_{\boldsymbol{\ell}}({\mathbf{x}_{\boldsymbol{\ell}}})\big|\bigg\}\bigg],
\end{multline*}
where $\mathcal{X}_{\boldsymbol{\ell}} = \prod_{j \in \boldsymbol{\ell}} \mathcal{X}_j$. This bound is loose because it is likely attained at an impossible configuration of the inputs. For instance, the loose bound could combine an age of 23 for the main effect with an age of 54 for the interaction term. Although this is problematic, it offers a tractable approximation of the bound by simplifying the computation.

The second approach introduces graphs to leverage dynamic programming in the computation of sharp bounds. The intuition is to optimise over multiple subsets of dependent covariates instead of over all covariates simultaneously. For $a \in \{\text{pre}, \text{post}\}$, let $\mathcal{G}^a = (\mathcal{V}^a, \mathcal{E}^a)$ denote a graph encoding pairwise relations among covariates, where $\mathcal{V}^a$ is the vertex set indexing the selected covariates and~$\mathcal{E}^a$ is the edge set representing pairwise interactions. Taking the union of the two vertex sets and the two edge sets,
\begin{align}
    \mathcal{V} = \mathcal{V}^{\text{pre}} \cup \mathcal{V}^{\text{post}}, \qquad \mathcal{E} = \mathcal{E}^{\text{pre}} \cup \mathcal{E}^{\text{post}},
\end{align}
yield the combined graph $\mathcal{G} = (\mathcal{V}, \mathcal{E})$. Let $\pi(\cdot)$ denote a function that 
partitions the vertices of an input graph into groups of connected vertices $\mathcal{B}$; for each group, denote $\mathcal{E}(\mathcal{B})$ the set of edges within it, and $\mathcal{X}_{\mathcal{B}} = \prod_{j \in \mathcal{B}} \mathcal{X}_j$ the domain of the associated covariates. Following this notation, the SADU is
\begin{multline}
\text{SADU} \big(\hat{\mathcal{F}}^{\text{post}}, \hat{\mathcal{F}}^{\text{post}}; \mathcal{X}\big) = g^{-1} \bigg[
\big|\hat{\beta}^{\text{post}}_{0} - \hat{\beta}^{\text{pre}}_{0}\big| + \\
\sum_{\mathcal{B} \in \pi(\mathcal{G})} \underset{\mathbf{x}_{\mathcal{B}} \in \mathcal{X}_{\mathcal{B}}}{\mathrm{sup}} \bigg\{ \sum_{j \in \mathcal{B}} \big|\hat{f}^{\text{post}}_{j}(x_j) - \hat{f}^{\text{pre}}_{j}(x_j)\big| + \sum_{\boldsymbol{\ell} \in \mathcal{E}(\mathcal{B})} \big|\hat{f}^{\text{post}}_{\boldsymbol{\ell}}(\mathbf{x}_{\boldsymbol{\ell}}) 
    - \hat{f}^{\text{pre}}_{\boldsymbol{\ell}}(\mathbf{x}_{\boldsymbol{\ell}})\big| \bigg\}\bigg], \label{eq:optim_sharp_bound_graph}
\end{multline}
where $\mathbf{x}_{\mathcal{B}} = (x_j)_{j \in \mathcal{B}}$ and $\mathbf{x}_{\boldsymbol{\ell}} = (x_j)_{j \in \boldsymbol{\ell}}$. This bound is sharp because it is attained at an achievable input configuration. Equation~\eqref{eq:optim_sharp_bound_graph} is a computationally convenient reformulation of Equation~\eqref{eq:optim_sharp_bound}. 

\section{Training and retraining methodology}
\label{app:train_retrain_metho}

We detail the training and retraining procedures for each model class. Recall that for all models and datasets, we use a log link function and we select the loss function according to the response: Poisson deviance for claim frequency, gamma deviance for claim severity, and Tweedie deviance for loss cost. 

For GLMs and GAMs, that do not perform interaction selection by default, interaction terms are not considered. EBMs use the FAST interaction detection algorithm, and XGBoost models inherently capture interactions because of their tree-based formulation.

For the choice of shape functions, GLMs are limited to linear effects. For GAMs, we use penalised thin plate regression splines because they are the optimal smoother of any given basis dimension \citep{wood2003thin}. EBMs, by definition, use bagged-boosted shallow decision trees.

EBM and XGBoost require the tuning of several hyperparameters. We set both learning rates (\texttt{learning\_rate}) to $0.01$. For EBMs, we perform a grid search to tune the number of initial boosting rounds heavily regularised for main effects (\texttt{smoothing\_rounds}), the number of interaction terms (specified as a proportion of the number of covariates) to be included (\texttt{interactions}), the maximum number of bins per feature for the main effects stage (\texttt{max\_bins}), and the maximum number of leaves allowed in each tree (\texttt{max\_leaves}), similarly to \citet{krupova2026explainableboostingmachinepredicting}. 

For XGBoost models, we set \texttt{min\_child\_weight}, the minimum number of observations needed in each node, to $40$, and both the subsample ratio of the training observations at each boosting iteration (\texttt{subsample}) and that of columns for each  depth level reached in a tree (\texttt{colsample\_bylevel}) to $0.75$. We perform a grid search to tune the maximum depth of a tree (\texttt{max\_depth}) and the number of trees (\texttt{iteration\_range}). All unmentioned hyperparameters are kept at their default values for both EBM and XGBoost models. In Tables~\ref{tab:ebm_hyperparam_grid}~to~\ref{tab:xgb_hyperparam_public}, we present the grids used and the resulting optimal hyperparameters for each model on each dataset.

We now describe our retraining procedure. Because the selected datasets do not have a temporal variable, we follow Algorithm~\ref{alg:data-split-variation} to construct the pre- and post-update training and test datasets. We set an update proportion of $\beta=0.3$ and a training proportion of $\alpha=0.7$. To properly select hyperparameters when needed, the pre- and post-update training datasets are further partitioned into non-validation and validation sets using the same $\alpha$ to define the non-validation proportion. We train all pre-update models using the pre-update training datasets and all post-update models using the post-update training datasets. Before generating predictions, models are globally rebalanced, multiplying their outputs by a scaling factor so that the sum of the fitted values over the training set matches the sum of the observed responses.

\begin{table}[h!]
\centering
\caption{Hyperparameter grid used to tune EBM models.}
\label{tab:ebm_hyperparam_grid}
\begin{tabular}{ll}
\toprule
Hyperparameter & Grid \\
\midrule
\texttt{smoothing\_rounds} & $\{100, 500, 1000\}$ \\
\texttt{interactions}      & $\{0.5, 0.75, 0.9\}$ \\
\texttt{max\_bins}         & $\{32, 1024, 4096\}$ \\
\texttt{max\_leaves}       & $\{2, 3, 4\}$        \\
\bottomrule
\end{tabular}
\end{table}

\begin{table}[h!]
\centering
\caption{Hyperparameter grid used to tune XGBoost models.}
\label{tab:xgb_hyperparam_grid}
\begin{tabular}{ll}
\toprule
Hyperparameter & Grid \\
\midrule
\texttt{max\_depth}       & $\{3, 5, 7\}$               \\
\texttt{iteration\_range} & $\{100, 300, \dots, 1900\}$ \\
\bottomrule
\end{tabular}
\end{table}

\begin{table}[h!]
\centering
\caption{Optimal hyperparameters for EBM models.}
\label{tab:ebm_hyperparam_public}
{\setlength{\tabcolsep}{4pt}
\begin{tabular}{ll rrrrrr}
\toprule
& & \multicolumn{2}{c}{\texttt{norauto}} & \multicolumn{2}{c}{\texttt{beMTPL97}} & \multicolumn{2}{c}{\texttt{pg15training}} \\
\cmidrule(lr){3-4} \cmidrule(lr){5-6} \cmidrule(lr){7-8}
Model & Hyperparameter & Pre & Post & Pre & Post & Pre & Post \\
\midrule
\multirow{1.5}{*}{Frequency}
  & \texttt{smoothing\_rounds} & 100  & 1000 & 500  & 500  & 100 & 100 \\
  & \texttt{interactions}      & 0.9  & 0.9  & 0.75 & 0.75 & 0.9 & 0.5 \\
  & \texttt{max\_bins}         & 32   & 32   & 32   & 1024 & 4096 & 1024 \\
  & \texttt{max\_leaves}       & 3    & 4    & 3    & 4    & 3 & 4 \\
\midrule
\multirow{1.5}{*}{Severity}
  & \texttt{smoothing\_rounds} & 500  & 500  & 500  & 100  & 1000 & 100 \\
  & \texttt{interactions}      & 0.9  & 0.75 & 0.5  & 0.9  & 0.9 & 0.75 \\
  & \texttt{max\_bins}         & 32   & 32   & 32   & 4096 & 32 & 4096 \\
  & \texttt{max\_leaves}       & 2    & 2    & 2    & 3    & 4 & 4 \\
\midrule
\multirow{1.5}{*}{Loss cost}
  & \texttt{smoothing\_rounds} & 100  & 100  & 1000 & 1000 & 100 & 100 \\
  & \texttt{interactions}      & 0.75 & 0.9  & 0.9  & 0.9  & 0.5 & 0.5 \\
  & \texttt{max\_bins}         & 32   & 32   & 1024 & 32   & 1024 & 1024 \\
  & \texttt{max\_leaves}       & 4    & 4    & 2    & 3    & 4 & 4 \\
\bottomrule
\end{tabular}}
\end{table}

\begin{table}[h!]
\centering
\caption{Optimal hyperparameters for XGBoost models.}
\label{tab:xgb_hyperparam_public}
{\setlength{\tabcolsep}{4pt}
\begin{tabular}{ll rrrrrr}
\toprule
& & \multicolumn{2}{c}{\texttt{norauto}} & \multicolumn{2}{c}{\texttt{beMTPL97}} & \multicolumn{2}{c}{\texttt{pg15training}} \\
\cmidrule(lr){3-4} \cmidrule(lr){5-6} \cmidrule(lr){7-8}
Model & Hyperparameter & Pre & Post & Pre & Post & Pre & Post \\
\midrule
\multirow{1.5}{*}{Frequency}
  & \texttt{max\_depth}       & 3    & 3    & 3    & 5    &  5 & 5 \\
  & \texttt{iteration\_range} & 1700 & 1700 & 1900 & 1900 & 1700 & 1900 \\
\midrule
\multirow{1.5}{*}{Severity}
  & \texttt{max\_depth}       & 3    & 3    & 3    & 3    & 3 & 3 \\
  & \texttt{iteration\_range} & 1500 & 1500 & 1100 & 1100 & 1100 & 1100 \\
\midrule
\multirow{1.5}{*}{Loss cost}
  & \texttt{max\_depth}       & 3    & 3    & 3    & 3    & 3 & 3 \\
  & \texttt{iteration\_range} & 700  & 700  & 700  & 900  & 900 & 1100 \\
\bottomrule
\end{tabular}}
\end{table}

\newpage
\section{Model-level results}
\label{app:model_level_results}

\begin{table}[h!]
    \centering
    \caption{Average rank of each metric over the three responses.}
    \label{tab:avg_ranks_all}
    \begin{tabular}{ll ccc}
    \toprule
    Dataset & Model
      & RMSS & MRD & Dev \\
    \midrule
    \multirow{1.5}{*}{\texttt{norauto}}
      & GLM     & 2.0          & \textbf{1.3} & \textbf{1.3} \\
      & EBM     & \textbf{1.0} & 2.0          & 2.3          \\
      & XGBoost & $-$          & 2.7          & 2.3          \\
    \midrule
    \multirow{1.5}{*}{\texttt{beMTPL97}}
      & GLM     & \textbf{1.0} & \textbf{1.0} & 3.3          \\
      & GAM     & 3.0          & 2.7          & 2.7          \\
      & EBM     & 2.0          & 3.0          & \textbf{2.0} \\
      & XGBoost & $-$          & 3.3          & \textbf{2.0} \\
    \midrule
    \multirow{1.5}{*}{\texttt{pg15training}}
      & GLM     & \textbf{1.0} & \textbf{1.3} & 3.3          \\
      & GAM     & 3.0          & 2.7          & 2.0          \\
      & EBM     & 2.0          & 4.0          & \textbf{1.7} \\
      & XGBoost & $-$          & 2.0          & 3.0          \\
    \bottomrule
    \end{tabular}
\end{table}

We present the average rank of each metric over the three responses for each combination of dataset--model (see Table~\ref{tab:avg_ranks_all}) and the detailed stability and performance results (see Table~\ref{tab:combined_all}). All displayed metrics are computed on the post-update test dataset. As \texttt{norauto} contains only categorical variables, the GAM reduces to a standard GLM; results are thus not reported for the GAM on this dataset.

\begin{table}[t!]
\footnotesize
    \centering
    \caption{Model stability and performance across datasets for all three responses.}
    \label{tab:combined_all}
    % \small
    \setlength{\tabcolsep}{4pt}
    \begin{tabular}{ll ccc ccc ccc}
    \toprule
    & & \multicolumn{3}{c}{Frequency} & \multicolumn{3}{c}{Severity} & \multicolumn{3}{c}{Loss cost} \\
    \cmidrule(lr){3-5} \cmidrule(lr){6-8} \cmidrule(lr){9-11}
    Dataset & Model & RMSS & MRD & Dev & RMSS & MRD & Dev & RMSS & MRD & Dev \\
    \midrule
    \multirow{1.5}{*}{\texttt{norauto}}
      & GLM     & 0.0424          & \textbf{0.0229} & \textbf{0.2696} & 0.0391          & 0.0241          & 1.0620          & 0.0830          & \textbf{0.0415} & \textbf{200.67} \\
      & EBM     & \textbf{0.0323} & 0.0241          & 0.2775          & \textbf{0.0230} & \textbf{0.0147} & \textbf{1.0596} & \textbf{0.0626} & 0.0523          & 206.78          \\
      & XGBoost & ---             & 0.0273          & 0.2698          & ---             & 0.0318          & 1.0623          & ---             & 0.0501          & 200.77          \\
    \midrule
    \multirow{1.5}{*}{\texttt{beMTPL97}}
      & GLM     & \textbf{0.0208} & \textbf{0.0167} & 0.5371          & \textbf{0.0650} & \textbf{0.0440} & 2.0530          & \textbf{0.0623} & \textbf{0.0442} & 75.29          \\
      & GAM     & 0.1592          & 0.0322          & 0.5368          & 0.0917          & 0.0640          & \textbf{2.0454} & 0.2131          & 0.0787          & 103.58         \\
      & EBM     & 0.1083          & 0.0611          & 0.5346          & 0.0718          & 0.0544          & 2.0469          & 0.1700          & 0.1271          & 75.26          \\
      & XGBoost & ---             & 0.0748          & \textbf{0.5327} & ---             & 0.0571          & 2.0568          & ---             & 0.0926          & \textbf{74.92} \\
    \midrule
    \multirow{1.5}{*}{\texttt{pg15training}}
      & GLM     & \textbf{0.0627} & \textbf{0.0409} & 0.5032          & \textbf{0.0501} & 0.0348          & 1.0399          & \textbf{0.0910} & \textbf{0.0642} & 47.13          \\
      & GAM     & 4.1650          & 0.0527          & 0.4915          & 0.7087          & 0.0358          & \textbf{1.0391} & 12.2137         & 0.0992          & 46.17          \\
      & EBM     & 0.1433          & 0.1059          & \textbf{0.4864} & 0.0623          & 0.0449          & 1.0420          & 0.1803          & 0.1433          & \textbf{46.00} \\
      & XGBoost & ---             & 0.0846          & 0.4911          & ---             & \textbf{0.0332} & 1.0495          & ---             & 0.0908          & 46.22          \\
    \bottomrule
    \end{tabular}
\end{table}

% \clearpage
\section{Other structurally unstable feature on \texttt{beMTPL97} with the claim frequency response}
\label{app:instability-driving-features}

\begin{figure}[h!]
\vspace{-12pt}
    \centering
    \makebox[\textwidth][c]{%
      \includegraphics[width = 0.945\textwidth]{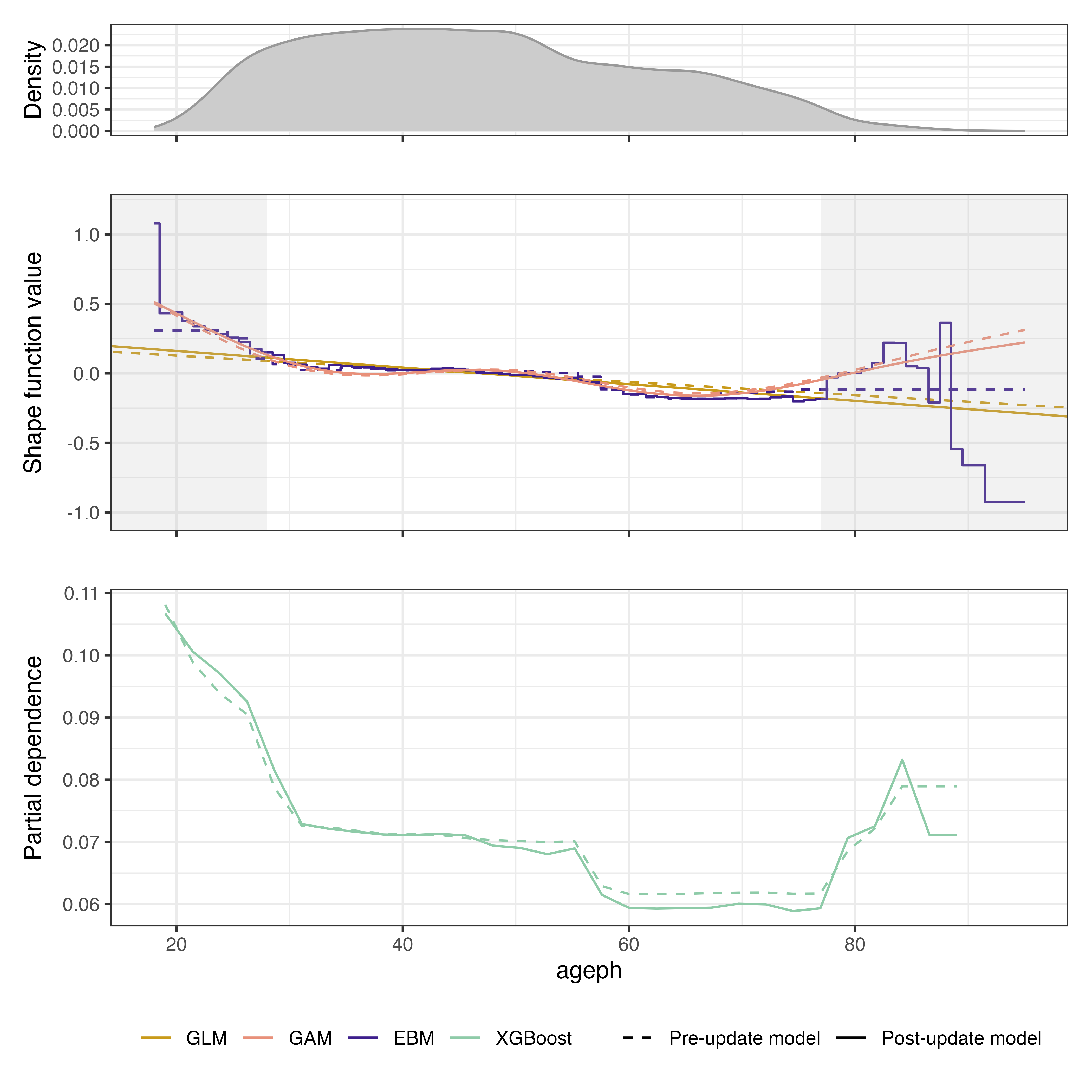}}
    \caption{Feature density (top), pre- and post-update shape functions from the three transparent models (middle), and pre- and post-update partial dependence plots from XGBoost (bottom) for \texttt{ageph} with the frequency in \texttt{beMTPL97}.}
    \label{fig:shape_functions_ageph}
\end{figure}

As illustrated in Figure~\ref{fig:shape_functions_ageph}, the instability of \texttt{ageph} is largely concentrated in the tails of the distribution (see shaded regions in the middle panel) for all three transparent models. For the GLM, \texttt{ageph} is the only feature with a noticeable effect on the MSS. The divergence of the pre- and post-update shape functions in the tails for GLM and EBM, and the agreement between the XGBoost partial dependence functions and the shape functions in terms of trend (compare middle and bottom panels), are both consistent with the remarks about Figures~\ref{fig:shape_functions_agec}~and~\ref{fig:shape_functions_lat}.

\section{Additional results on stabilisation strategies}
\label{app:add_res_stab_strat}

In Tables~\ref{tab:seq_avg_ranks_all}~and~\ref{tab:seq_combined_all}, we first present the model-level results for the GLM, EBM, and XGBoost models under sequential modelling for all datasets, in the same format as \appref{app:model_level_results}. We compute all the metrics on the post-update test datasets. In Table~\ref{tab:combined_all_fixed}, we present the model-level results for EBM and XGBoost models with fixed hyperparameters for all combinations of dataset--response. In Table~\ref{tab:combined_all_fixedfeat}, we present the model-level results for EBMs with fixed interaction terms, with and without fixed hyperparameters, for all datasets. In Table~\ref{tab:ebm_same_int_rmss_mrd_deviance}, we show the variations for all metrics between EBM with and without fixed interaction terms.

\begin{table}[H]
\vspace{-12pt}
    \centering
    \caption{Average rank of each metric over the three responses under sequential modelling.}
    \label{tab:seq_avg_ranks_all}
    \begin{tabular}{ll ccc}
    \toprule
    Dataset & Model
      & RMSS & MRD & Dev \\
    \midrule
    \multirow{1.5}{*}{\texttt{norauto}}
      & GLM     & 2.0          & 3.0          & \textbf{1.0} \\
      & EBM     & \textbf{1.0} & \textbf{1.0} & 2.7          \\
      & XGBoost & $-$          & 2.0          & 2.3          \\
    \midrule
    \multirow{1.5}{*}{\texttt{beMTPL97}}
      & GLM     & \textbf{1.3} & 2.0          & 2.0          \\
      & EBM     & 1.7          & 2.3          & 2.3          \\
      & XGBoost & $-$          & \textbf{1.7} & \textbf{1.7} \\
    \midrule
    \multirow{1.5}{*}{\texttt{pg15training}}
      & GLM     & \textbf{1.3} & 3.0          & 2.3          \\
      & EBM     & 1.7          & \textbf{1.0} & \textbf{1.7} \\
      & XGBoost & $-$          & 2.0          & 2.0          \\
    \bottomrule
    \end{tabular}
\end{table}

\begin{table}[t!]
\footnotesize
    \centering
    \caption{Model stability and performance across datasets for all three responses under sequential modelling.}
    \label{tab:seq_combined_all}
    % \small
    \setlength{\tabcolsep}{4pt}
    \begin{tabular}{ll ccc ccc ccc}
    \toprule
    & & \multicolumn{3}{c}{Frequency} & \multicolumn{3}{c}{Severity} & \multicolumn{3}{c}{Loss cost} \\
    \cmidrule(lr){3-5} \cmidrule(lr){6-8} \cmidrule(lr){9-11}
    Dataset & Model & RMSS & MRD & Dev & RMSS & MRD & Dev & RMSS & MRD & Dev \\
    \midrule
    \multirow{1.5}{*}{\texttt{norauto}}
      & GLM     & 0.0424          & 0.0229          & \textbf{0.2696} & 0.0391          & 0.0241          & \textbf{1.0620} & 0.0830          & 0.0415          & \textbf{200.67} \\
      & EBM     & \textbf{0.0101} & \textbf{0.0066} & 0.2776          & \textbf{0.0238} & \textbf{0.0049} & 1.0623          & \textbf{0.0285} & \textbf{0.0270} & 206.91          \\
      & XGBoost & ---             & 0.0187          & 0.2699          & ---             & 0.0130          & 1.0667          & ---             & 0.0345          & 200.87          \\
    \midrule
    \multirow{1.5}{*}{\texttt{beMTPL97}}
      & GLM     & \textbf{0.0208} & \textbf{0.0167} & 0.5371          & 0.0650          & 0.0440          & \textbf{2.0530} & \textbf{0.0623} & 0.0442          & 75.29          \\
      & EBM     & 0.0518          & 0.0241          & 0.5352          & \textbf{0.0211} & \textbf{0.0192} & 2.0631          & 0.0693          & 0.0468          & 76.88          \\
      & XGBoost & ---             & 0.0206          & \textbf{0.5326} & ---             & 0.0284          & 2.0695          & ---             & \textbf{0.0369} & \textbf{75.11} \\
    \midrule
    \multirow{1.5}{*}{\texttt{pg15training}}
      & GLM     & \textbf{0.0627} & 0.0409          & 0.5032          & 0.0501          & 0.0348          & \textbf{1.0399} & \textbf{0.0910} & 0.0642          & 47.13          \\
      & EBM     & 0.0628          & \textbf{0.0200} & \textbf{0.4897} & \textbf{0.0240} & \textbf{0.0134} & 1.0427          & 0.0911          & \textbf{0.0166} & 46.35          \\
      & XGBoost & ---             & 0.0204          & 0.4926          & ---             & 0.0211          & 1.0481          & ---             & 0.0240          & \textbf{46.32} \\
    \bottomrule
    \end{tabular}
\end{table}

\begin{table}[t!]
\footnotesize
    \centering
    \caption{Model stability and performance across datasets for all three responses using fixed hyperparameters.}
    \label{tab:combined_all_fixed}
    \setlength{\tabcolsep}{4pt}
    \begin{tabular}{ll ccc ccc ccc}
    \toprule
    & & \multicolumn{3}{c}{Frequency} & \multicolumn{3}{c}{Severity} & \multicolumn{3}{c}{Loss cost} \\
    \cmidrule(lr){3-5} \cmidrule(lr){6-8} \cmidrule(lr){9-11}
    Dataset & Model & RMSS & MRD & Dev & RMSS & MRD & Dev & RMSS & MRD & Dev \\
    \midrule
    \multirow{1.5}{*}{\texttt{norauto}}
      & EBM     & 0.0294 & \textbf{0.0230} & 0.2775          & 0.0236 & \textbf{0.0142} & \textbf{1.0598} & 0.0623 & 0.0519          & 206.81          \\
      & XGBoost & ---    & 0.0268          & \textbf{0.2698} & ---    & 0.0318          & 1.0626          & ---    & \textbf{0.0476} & \textbf{200.76} \\
    \midrule
    \multirow{1.5}{*}{\texttt{beMTPL97}}
      & EBM     & 0.0720 & 0.0521          & 0.5352          & 0.0642 & \textbf{0.0508} & \textbf{2.0480} & 0.0996 & \textbf{0.0793} & 76.12          \\
      & XGBoost & ---    & \textbf{0.0400} & \textbf{0.5326} & ---    & 0.0569          & 2.0587          & ---    & 0.0837          & \textbf{74.90} \\
    \midrule
    \multirow{1.5}{*}{\texttt{pg15training}}
      & EBM     & 0.1311 & 0.0893          & \textbf{0.4876} & 0.0448 & 0.0357          & \textbf{1.0421} & 0.1732 & 0.1384          & \textbf{46.00} \\
      & XGBoost & ---    & \textbf{0.0824} & 0.4918          & ---    & \textbf{0.0328} & 1.0489          & ---    & \textbf{0.0911} & 46.35          \\
    \bottomrule
    \end{tabular}
\end{table}

\begin{table}[t!]
    \footnotesize
    \centering
    \caption{Model stability and performance across datasets for all three responses using EBMs with fixed features.}
    \label{tab:combined_all_fixedfeat}
    \setlength{\tabcolsep}{4pt}
    \begin{tabular}{ll ccc ccc ccc}
    \toprule
    & & \multicolumn{3}{c}{Frequency} & \multicolumn{3}{c}{Severity} & \multicolumn{3}{c}{Loss cost} \\
    \cmidrule(lr){3-5} \cmidrule(lr){6-8} \cmidrule(lr){9-11}
    Dataset & \makecell[lb]{Fixed\\hyperparameters} & RMSS & MRD & Dev & RMSS & MRD & Dev & RMSS & MRD & Dev \\
    \midrule
    \multirow{1.5}{*}{\texttt{norauto}}
      & No   & 0.0329          & 0.0241          & 0.2775          & \textbf{0.0230} & \textbf{0.0145} & 1.0601          & 0.0619 & 0.0514 & 206.79 \\
      & Yes  & \textbf{0.0291} & \textbf{0.0230} & 0.2775          & 0.0236          & 0.0146          & 1.0601          & 0.0619 & 0.0514 & 206.79 \\
    \midrule
    \multirow{1.5}{*}{\texttt{beMTPL97}}
      & No   & 0.1230          & 0.0569          & \textbf{0.5348} & 0.0711          & 0.0530          & \textbf{2.0473} & 0.1710          & 0.1411          & \textbf{75.16} \\
      & Yes  & \textbf{0.0601} & \textbf{0.0432} & 0.5351          & \textbf{0.0642} & \textbf{0.0509} & 2.0480          & \textbf{0.0924} & \textbf{0.0725} & 75.99          \\
    \midrule
    \multirow{1.5}{*}{\texttt{pg15training}}
      & No   & 0.1404          & 0.0875          & \textbf{0.4870} & \textbf{0.0415} & \textbf{0.0345} & 1.0418          & 0.1664 & 0.1313 & 46.01 \\
      & Yes  & \textbf{0.1238} & \textbf{0.0827} & 0.4876          & 0.0430          & 0.0348          & \textbf{1.0416} & 0.1664 & 0.1313 & 46.01 \\
    \bottomrule
    \end{tabular}
\end{table}

\begin{table}[t!]
\centering
\footnotesize
\caption{Variations (in \%) of each metric for EBMs when fixing interaction terms.}
\label{tab:ebm_same_int_rmss_mrd_deviance}
{\setlength{\tabcolsep}{2pt}
\begin{tabular}{lrrrrrrrrr}
\toprule
& \multicolumn{3}{c}{\texttt{norauto}} & \multicolumn{3}{c}{\texttt{beMTPL97}} & \multicolumn{3}{c}{\texttt{pg15training}} \\
\cmidrule(lr){2-4} \cmidrule(lr){5-7} \cmidrule(lr){8-10}
& Frequency & Severity & Loss cost & Frequency & Severity & Loss cost & Frequency & Severity & Loss cost \\
\midrule
RMSS & $+$1.8 & $-$0.1 & $-$1.1 & $+$13.7 & $-$0.9 & $+$0.6 & $-$2.0 & $-$33.5 & $-$7.7 \\
MRD & $+$0.1 & $-$1.0 & $-$1.8 & $-$6.9 & $-$2.4 & $+$11.0 & $-$17.4 & $-$23.1 & $-$8.4 \\
Dev & $+$0.001 & $+$0.049 & $+$0.002 & $+$0.025 & $+$0.017 & $-$0.132 & $+$0.130 & $-$0.015 & $+$0.018 \\
\bottomrule
\end{tabular}}
\end{table}

\section{GLM and sequential modelling}
\label{app:glm_seq_mod}

To ensure consistency with our case study of Section~\ref{sec:case_study}, we assume that GLMs are always trained on centred covariates. Using the pre-update predictions as fixed offsets in a GLM changes the interpretation of the coefficients in the post-update model. Instead of describing the effect of the feature on the predictions, they describe the effect of the feature on the residuals. We call them the update coefficients because they are not sufficient to represent their associated feature's shape function on their own; for a categorical main effect $j \in \mathcal{J}$, the post-update shape function is the sum of the pre-update coefficient $\hat{\beta}^{\text{pre}}_{j}$ and the update coefficient $\hat{\beta}^{\text{up}}_{j}$. For a numerical main effect $j$, the pre-update shape function is
\begin{align*}
    \hat{f}^{\text{pre}}_{j}(x_{ij}) = \hat{\beta}^{\text{pre}}_{j} (x_{ij} - \Bar{x}^{\text{pre}}_{j}).
\end{align*}
The post-update shape function is
\begin{align*}
    \hat{f}^{\text{post}}_{j}(x_{ij}) &= \hat{\beta}^{\text{pre}}_{j} (x_{ij} - \Bar{x}^{\text{pre}}_{j}) + \hat{\beta}^{\text{up}}_{j} (x_{ij} - \Bar{x}^{\text{post}}_{j})\\
    &= \hat{\beta}^{\text{post}}_{j} x_{ij} - (\hat{\beta}^{\text{pre}}_{j}\Bar{x}^{\text{pre}}_{j} + \hat{\beta}^{\text{up}}_{j}\Bar{x}^{\text{post}}_{j}),
\end{align*} 
where the post-update coefficient 
\begin{align}
    \hat{\beta}^{\text{post}}_{j} = \hat{\beta}^{\text{pre}}_{j} + \hat{\beta}^{\text{up}}_{j} \label{eq:beta_post}
\end{align}
is the coefficient we would obtain without sequential modelling. Because the post-update shape function cannot be written as that of a GLM without sequential modelling, that is,
\begin{align*}
    \hat{f}^{\text{post}}_{j}(x_{ij})
    &= \hat{\beta}^{\text{post}}_{j} (x_{ij} - \Bar{x}^{\text{post}}_{j})\\
    &= \hat{\beta}^{\text{post}}_{j} x_{ij} - \hat{\beta}^{\text{post}}_{j}\Bar{x}^{\text{post}}_{j},
\end{align*}
the intercept requires an adjustment so that Equation~\eqref{eq:beta_post} holds true for all main effects. The same adjustment applies to interactions involving at least one numerical feature.

Put differently, the intercept adjustment accounts for the vertical shift of post-update shape functions that involve numerical effects, relative to the post-update shape function of a GLM without sequential modelling; for numerical main effects, these vertical shifts are 
$$
\hat{\beta}^{\text{pre}}_{j}\Bar{x}^{\text{pre}}_{j} + \hat{\beta}^{\text{up}}_{j}\Bar{x}^{\text{post}}_{j} - \hat{\beta}^{\text{post}}_{j}\Bar{x}^{\text{post}}_{j}.
$$
Both structural and predictive stability are affected by sequential modelling.

Rebalancing the pre-update predictions before using them as fixed offsets modifies the intercept of the post-update model when compared to a GLM estimated without sequential modelling. Hence, 
$$
\hat{\beta}^{\text{post}}_{0} \neq \hat{\beta}^{\text{pre}}_{0} + \hat{\beta}^{\text{up}}_{0},
$$
because rebalancing is equivalent to adding the logarithm of a scaling factor to the intercept. The update intercept $\hat{\beta}^{\text{up}}_{0}$ is adjusted to account for the change in scale of the offsets. Rebalancing the post-update predictions eliminates the impact of the update intercept change, such that
sequential modelling has no impact on the predictive stability of a rebalanced GLM. The structural stability, however, is still affected by the vertical shift of the post-update shape functions.

By comparing the results in \appref{app:model_level_results} with those of \appref{app:add_res_stab_strat}, we see that sequential modelling has no impact on the predictive stability or performance of the GLM. For structural stability, this is not the case, though the difference is so negligible that it is not apparent even at four decimal places. This is expected with large datasets because if $\Bar{x}^{\text{pre}}_{j} \approx \Bar{x}^{\text{post}}_{j}$ for all numerical main effects $j$, then the vertical shifts of the post-update shape functions are 
\begin{align*}
    \hat{\beta}^{\text{pre}}_{j}\Bar{x}^{\text{pre}}_{j} + \hat{\beta}^{\text{up}}_{j}\Bar{x}^{\text{post}}_{j} - \hat{\beta}^{\text{post}}_{j}\Bar{x}^{\text{post}}_{j} &\approx \hat{\beta}^{\text{pre}}_{j}\Bar{x}^{\text{post}}_{j} + \hat{\beta}^{\text{up}}_{j}\Bar{x}^{\text{post}}_{j} - \hat{\beta}^{\text{post}}_{j}\Bar{x}^{\text{post}}_{j} 
    \\ &= \hat{\beta}^{\text{post}}_{j}\Bar{x}^{\text{post}}_{j} - \hat{\beta}^{\text{post}}_{j}\Bar{x}^{\text{post}}_{j} 
    \\ &= 0.
\end{align*}
Sequential modelling barely affects the structural stability of the GLM.

\end{document}